\documentclass[pdflatex,sn-mathphys-num]{sn-jnl}

\usepackage{amssymb}
\usepackage{amsmath}

\usepackage{lipsum}
\usepackage{graphicx}
\usepackage{lscape}
\usepackage{subcaption}
\usepackage{color, colortbl}
\definecolor{beige}{rgb}{0.96, 0.96, 0.86}
 \definecolor{blanchedalmond}{rgb}{1.0, 0.92, 0.8}
\definecolor{bubbles}{rgb}{0.91, 1.0, 1.0}
\definecolor{eggshell}{rgb}{0.94, 0.92, 0.84}
\definecolor{ghostwhite}{rgb}{0.97, 0.97, 1.0}
\definecolor{lightCyan}{rgb}{0.80, 0.80, 0.80}
\definecolor{lightCyan1}{rgb}{0.85, 0.85, 0.85}

\theoremstyle{thmstyleone}%
\theoremstyle{thmstyletwo}%

\theoremstyle{thmstylethree}%

\begin{document}

\title[Article Title]{$OpenCML$: End-to-End Framework of Open-world Machine Learning to Learn Unknown Classes Incrementally}

\author*[1]{\fnm{Jitendra} \sur{Parmar}}\email{jitendra.parmar@juetguna.in}

\author[2]{\fnm{Praveen Singh} \sur{Thakur}}\email{praveen.thakur@nmims.edu}
\equalcont{These authors contributed equally to this work.}

\affil*[1]{\orgdiv{Computer Science and Engineering}, \orgname{Jaypee University of Engineering and Technology}, \orgaddress{\street{Agra-Mumbai Hwy}, \city{Guna}, \postcode{473226}, \state{Madhya Pradesh}, \country{India}}}

\affil[2]{\orgdiv{Computer  Engineering}, \orgname{SVKM'S NMIMS STME}, \orgaddress{\street{Super Corridor Rd}, \city{Indore}, \postcode{453112}, \state{Madhya Pradesh}, \country{India}}}

\abstract{
Open-world machine learning is an emerging technique in artificial intelligence, where conventional machine learning models often follow closed-world assumptions, which can hinder their ability to retain previously learned knowledge for future tasks. However, automated intelligence systems must learn about novel classes and previously known tasks. The proposed model offers novel learning classes in an open and continuous learning environment. It consists of two different but connected tasks. First, it discovers unknown classes in the data and creates novel classes; next, it learns how to perform class incrementally for each new class. Together, they enable continual learning, allowing the system to expand its understanding of the data and improve over time. The proposed model also outperformed existing approaches in open-world learning. Furthermore, it demonstrated strong performance in continuous learning, achieving a highest average accuracy of 82.54\% over four iterations and a minimum accuracy of 65.87\%.
}
\maketitle






\section{Introduction}
Human beings learn from childhood by experiencing and analysing things; they build a knowledge base (memories). It is an incremental learning process; generally, they do not forget this learning easily and utilise it for a long time in their future tasks. This phenomenon of incremental learning reduces effort, as there is no need to repeat similar tasks. The classical machine learning technique does not follow incremental learning techniques; for this reason, it must repeatedly relearn old tasks, which decreases system performance. It is challenging to develop classical or traditional machine learning models that can learn incrementally, as humans do. Here, in the case of traditional machine learning, we have to train the machine every time it encounters unknown data.
This type of data is referred to as open-world data, which does not appear during training. In contrast, classical machine learning uses a closed-world assumption, where each testing data instance is available during the training.

Let us consider a dataset of 10,000 labelled text inputs, where each input is assigned to one of five intents: Transfer, Distance, Rewards Balance, Travel, and Utility. The model is trained on this dataset to classify the intents of new text inputs.
However, new intents may become apparent over time, missing from the initial data set. For example, users might start questioning regional events or cafe requests. We can use incremental class learning to update the model with new labelled data when available( to manage these unknown intents). 
For example, if we receive 100 new inputs that belong to an unknown intent, such as "events," we can provide these new data attributes to the model and modify its parameters accordingly. In this way, the model can adapt these unknown intentions without retraining the entire model for all data; also, it will not forget the previous knowledge.
By merging incremental class learning with OWML, the model can learn continuously and enhance its ability to determine unknown intents and classify data more accurately. It can benefit domains and applications where unknown data is continuously growing and varying, such as chatbots or voice assistants.

The literature contains numerous attempts to address the issue of open-world machine learning. learning~\cite{mccloskey1989catastrophic,kirkpatrick2017overcoming,fei2016breaking,shu2017doc,prakhya2017open,lin2019post,vedula2019towards,vedula2020automatic}. Similarly,  for class incremental learning~\cite{javed2019meta,gallardo2021self,mehta2021empirical,wu2022class,madaan2021representational,purushwalkam2022challenges,fini2022self,cha2021co2l,robbins1951stochastic,kiefer1952stochastic,bottou2018optimization,lopez2017gradient_gem,chaudhry2018efficient_agem,tang2021layerwise,riemer2018learning,kirkpatrick2017overcoming,schwarz2018progress,zenke2017continual_si,aljundi2018memory_mas,chaudhry2018riemannian_rwalk,benzing2022unifying,riemer2018learning,chaudhry2019tiny,lopez2017gradient_gem,rebuffi2017icarl,caccia2020online,belouadah2019il2m,gong2022continual,ebrahimi2020remembering,saha2020gradient} Among them, perhaps the most effective strategy is to keep a memory buffer that stores part of the previous knowledge for the rehearsal~\cite{Robins93a,Robins95}.Therefore, it needs a model to effectively integrate novel concepts without omitting the prior information, also known as the stability-plasticity dilemma. 
Excess plasticity usually yields a significant performance degradation of the old classes, which is referred to as catastrophic forgetting~\cite{FrenchC02}.  Although samples are saved in memory, they have many challenges in CML, which can be understood from the representation and classifier learning views~\cite{serra2018overcoming}. Representation learning refers to the process of learning data representations. Due to limited memory, classifier learning often requires improving the class imbalance between prior and novel classes. A restricted memory size typically leads to an imbalance between prior and novel classes.

To the best of our knowledge, no framework exists that can integrate both open-world learning and incremental class learning for text classification. In this paper, we propose OpenCML, an end-to-end framework for open-world machine learning that learns unknown classes incrementally, addressing both open-world learning and incremental class learning. Specifically, we address the following research questions using the proposed OpenCML framework.

\begin{itemize}
    \item $RQ1$: How can we deploy and scale CIL techniques in real-world applications in OWML?
    \item $RQ2$: How can we develop cost-efficient memory management and storage techniques for CIL in OWML?
    \item $RQ3$: How can we design methods to recognize and mitigate catastrophic forgetting in CML in OWML?
    \item $RQ4$: How can we integrate previous knowledge to enhance CIL in OWML?
\end{itemize}

The proposed framework can identify unknown instances in the test data using a Convolutional Neural Network (CNN)- based model, which employs a 1-vs-rest approach to distinguish instances from existing ones. From known classes, we used examples to create an exemplar. The instances that our model rejects are further used to create novel classes. We applied Balanced Iterative Reducing and Clustering using Hierarchies (BIRCH) to identify the optimal clusters among the unknown instances. After forming clusters of unknown instances, the singleRank keyword extraction technique is used to generate new labels for the newly formed classes. We employed a custom loss function to prevent forgetting while learning classes incrementally.

\begin{table}[]
\scriptsize
\caption{List of variables used in the methodology and experiments}
\begin{tabular}{ll|ll}
\hline
Notation  & Meaning                 & Notation          & Meaning                                 \\ \hline
$d_{im}$  & World Vector            & $T_h$             & Threshold                               \\
$F_m$     & Feature Map             & $L(\theta)$ & Custom Loss      \\
$f$       & Non-leaner Features     & $L_{Ds}(\theta)$ & Distillation loss  \\
$b$       & Bias                    & $L_{Ce}(\theta)$ & cross-entropy loss \\
$S_{cn}$  & Number of known Classes & $g_i$ & Ground Truth                      \\
$I$       & Indicator Function      & $T$               & distillation parameter                        \\
$I_c$     & Number of Instances     & $cl$ & Classification Layer \\
$C_f$     & Cluster Features        & $M$ &Exemplar Memory   \\
$N_d$     & Data Points             &    \\
$L_s$     & Linear Sum              &  \\
$S_s$     & Squared Sum             &     \\
$\beta_f$ & Branching Factor        &   \\
          &                         &                        \\
          &                         &                         \\ \hline
\end{tabular}
\label{tab:my-table}
\end{table}

\section{Literature Reviews}
Continual machine learning or lifelong machine learning was first introduced in the late $90$'s~\cite {thrun1995learning}. In the last decade, continuous machine learning has drawn significant attention in the deep learning and Natural Language Processing (NLP) communities, and it is also known as lifelong machine learning~\cite{chen2018lifelong}. The issue is that when a neural network is used to memorize a series of tasks, remembering the subsequent tasks may hinder the execution of the models learned for the primary tasks. However, the human brain retains an extraordinary proficiency to perform various assignments simultaneously without negatively hindering one another. Continual learning algorithms attempt to perform this identical capacity for neural networks and to solve the catastrophic forgetting 
problem~\cite{mccloskey1989catastrophic,kirkpatrick2017overcoming}.
\subsection{Open-world machine learning}
Traditional machine learning approaches have produced promising results for decades in all domains of data analysis. However,  it has some limitations~\cite{bottou2014machine,burkart2021survey,kotsiantis2007supervised}; it works with isolated data and learns without utilizing prior knowledge. The trained model can only function with the input instances for which identical samples have been used for training purposes.

In~\cite{fei2016breaking}, the authors presented a space learning technique, the Centre-based Similarity (CBS), to determine text in the open world. Each feature in the mean of the positive class and the feature vector of the document are transformed separately by CBS in the document space vector. In~\cite{shu2017doc}, the authors introduced Deep Open Classification (DOC) as a method for identifying unclassified classes that may not be available in existing training classes. The classifier is designed to accurately classify both documents belonging to the known training classes and those that are unknown. This approach is known as open classification. Its multi-layer architecture is based on CNN. In~\cite{prakhya2017open}, the authors introduced another technique based on convolutional neural networks that incorporates feature extraction methods. It involves converting the document into a vector using the Word2Vec approach to extract features and calculate the cosine similarity between the entire document vector and the computed document vector using a naïve approach.

In~\cite{lin2019post}, the authors presented a softmax-based model to determine the profound novelty and detection of novel instances. It is called Softmax and Deep Novelty (SMDN). It utilises a Softmax and Local Outlier Factor (LOF) approach to identify new instances and can be implemented with various models without changing their architecture. In~\cite{vedula2019towards}, the authors introduced a two-phase mechanism model that predicts the statement's intent and then tags it in the input statement for open intent detection. The model comprises a Bidirectional Long Short-Term Memory (BiLSTM) and a Conditional Random Field (CRF) that uses adversarial training to enhance its robustness and execution across various domains. It can automatically detect a user's intent in natural language without requiring prior knowledge. The methodology begins with identifying any pre-existing open intents, which are then labelled with corresponding actions and objectives in the input words. If no actions or objectives are associated with a noticed intent, it is labelled as "none".

In~\cite{vedula2020automatic},  the authors presented  ADVIN, the automatic discovery of novel intents and domains; it can discover novel domains and intents from anonymous data. ADVIN operates in three phases: first, the identification of unknown domains; second,  knowledge transfer; and finally, tagging of intents to their affiliated novel domains. It utilises BERT and multi-class classifiers to identify unknown intents. The DOC is used for determining unknown intents with hierarchical clustering to determine unknown classes of intents.

\subsection{Continual machine learning}
To effectively handle the complexities of the natural world, an intelligent system must continuously acquire, revise, retain, and apply knowledge to its existence. This capability, comprehended as continual learning, provides a basis for artificial intelligence systems to design themselves adaptively. The ability to employ continuous learning, known as continual learning, functions as a foundation for artificial intelligence systems to tailor their designs dynamically. Overall, the concept of continuous learning is typically associated with a phenomenon known as catastrophic forgetting, whereby the acquisition of new or distinct knowledge often results in a notable decrease in performance for previously learned tasks. Moreover, various developments have emerged over the last decade that have expanded the understanding and application of continuous learning.

There are numerous approaches to achieving continual learning, such as  Representation, Regularisation, Optimisation, and  Replay-based approaches.

\subsubsection{Representation-based CML}
 In addition to earlier research on acquiring sparse representations via meta-training \cite{javed2019meta}, present researchers have attempted to integrate the edges of self-supervised learning \cite{gallardo2021self}, and large-scale pre-training \cite{mehta2021empirical,wu2022class} to enhance the representations in initialisation and  CML. These two approaches are nearly coupled since the pre-training data is usually of a massive portion and without detailed labels. At the same time, the execution of self-supervised learning is primarily evaluated through the fine-tuning of downstream tasks. The pre-training needs unsupervised learning or self-supervised learning to process extensive portions of data without detailed labels.
The approach is designed to perform self-supervised learning, primarily using the contrastive loss for CML, which is generally based on the idea of contrastive learning. Regarding those self-supervised representations that are crucial to mitigating catastrophic forgetting, in~\cite{madaan2021representational}, the authors presented a Lifelong Unsupervised Mixup that achieves further advancements by interpolating between examples of the old and novel tasks. 

In~\cite{purushwalkam2022challenges}, the authors presented a Minimum redundancy-based approach that further encourages the variousness of knowledge replay by de-correlating the accumulated previous training examples. In~\cite{fini2022self}, the authors presented a practical and straightforward framework for Continual Self-Supervised Learning. The framework transforms the self-supervised loss into a compression technique by mapping the current representation phase to its prior phase, which aims to enhance the representations of Self-Supervised Learning models in a continual learning scenario. In~\cite{cha2021co2l}, the authors presented  Contrastive Continual Learning; this approach utilises a contrastive loss to preserve individual tasks and a self-supervised loss to filter information between the previous and present examples.

\subsubsection{Optimization-based CML}
Optimisation-based CML is a paradigm for training ML models on sequential data that constantly arrives over time. The model is updated iteratively by utilising stochastic gradient descent (SGD) or similar optimisation algorithms~\cite{robbins1951stochastic,kiefer1952stochastic,bottou2018optimization}. At each iteration, the model is trained on a tiny subset of the sequential data, and the model's parameters are updated based on the error between the model's predictions and the actual weights of the data.

Some optimization methods also use replay-based approaches, such as  Gradient episodic memory (GEM), Averaged-GEM, Layer-wise optimization by gradient decomposition, and Maximizing Transfer and Minimizing Interference~\cite{lopez2017gradient_gem, chaudhry2018efficient_agem, tang2021layerwise, riemer2018learning}. These techniques ensure the prior input and gradient space conservation via previous training examples.

\subsubsection{Regularization-based CML}
Regularisation-based CML algorithms generally employ two methods: first, regularisation, which enables the model to retain previous learning while learning new tasks, and second, distillation, which utilises prior models as trainers to guide the learning of new models.
Weight regularisation is utilised to regulate the variation of network parameters selectively. 

In~\cite{kirkpatrick2017overcoming}The authors proposed elastic weight consolidation (EWC),  a solution to overcome catastrophic forgetting in neural networks. The algorithm decreases the learning rate on specific weights according to their importance in earlier learned tasks. The significance of this method is evaluated in both supervised learning and reinforcement learning contexts, where numerous tasks can be learned sequentially without interfering with prior learning. 

In~\cite{schwarz2018progress}, the authors proposed a method that utilises a knowledge base (KB) to train a competent model to solve previously encountered concerns, which is related to an active column used to learn the current task efficiently. After learning a new task, the active column is refined into the KB, protecting any earlier obtained skills. This active learning process, followed by compression, needs no architecture development, access to or accumulation of earlier data, and no specific parameters related to the task. 
It is the online version of EWC. It is based on the Fisher information matrix(FIM); the FIM is revised recursively without requiring access to the task label. Some other weight regularisation methods are  there, such as synaptic intelligence (SI), Memory aware synapses, Riemannian walk for incremental learning,  FIM~\cite{zenke2017continual_si, aljundi2018memory_mas, chaudhry2018riemannian_rwalk, benzing2022unifying}

\subsubsection{Replay-based CML}
Replay-based CML works by accumulating and replaying earlier data examples or pieces of knowledge to the model during training. The model is trained on both recent data and a subset of previously known data, which prevents catastrophic forgetting of earlier learned knowledge. There are many techniques to execute the replay-based CML, such as   Reservoir Sampling~\cite{riemer2018learning,chaudhry2019tiny}, 
 which randomly accumulates a specified number of training examples from each input set. Second, Ring Buffer~\cite{lopez2017gradient_gem} further provides an equivalent number of previous training examples randomly selected per class. Third,  Mean-of-Feature~\cite{rebuffi2017icarl} specifies an equivalent number of previous training examples that are most comparable to the characteristic mean of the individual class. 

 The concept of Online Continual Compression involves the concurrent learning of compression processes and the storage of representative data from a non-independent and identically distributed (non-i.i.d.) data stream. To overcome this problem, in~\cite{caccia2020online}, the authors suggested adaptive quantisation modules that enhance continual online compression and preserve compact data for replay. 
 Furthermore, the authors propose various methods to maintain prior learning, including supplementary knowledge with class statistics and minimal storage requirements. The methods employed are Class Incremental Learning with Dual Memory, Syntax-Aware Memory Network, Explanations that Reduce Catastrophic Forgetting, and Gradient Projection Memory for Continual Learning~\cite{belouadah2019il2m, gong2022continual, ebrahimi2020remembering, saha2020gradient}.

\begin{table}[]
\centering
\caption{Summarised Literature Surveys}
\label{SLS}
\begin{tabular}{lll|lll}
\hline
Author                             & OTC     & CML     & Author                               & OTC & CML     \\ \hline
~\cite{robbins1951stochastic}      & -       & $\surd$ & ~\cite{rebuffi2017icarl}             & -   & $\surd$ \\
~\cite{kiefer1952stochastic}       & -       & $\surd$ & ~\cite{bottou2018optimization}       & -   & $\surd$ \\
~\cite{fei2016breaking}            & $\surd$ & -       & ~\cite{chaudhry2018efficient_agem}   & -   & $\surd$ \\
~\cite{shu2017doc}                 & $\surd$ & -       & ~\cite{riemer2018learning}           & -   & $\surd$ \\
~\cite{prakhya2017open}            & $\surd$ & -       & ~\cite{schwarz2018progress}          & -   & $\surd$ \\
~\cite{lopez2017gradient_gem}      & -       & $\surd$ & ~\cite{aljundi2018memory_mas}        & -   & $\surd$ \\
~\cite{kirkpatrick2017overcoming}  & -       & $\surd$ & ~\cite{chaudhry2018riemannian_rwalk} & -   & $\surd$ \\
~\cite{zenke2017continual_si}      & -       & $\surd$ & ~\cite{riemer2018learning}           & -   & $\surd$ \\
~\cite{lin2019post}                & $\surd$ & -       & ~\cite{chaudhry2019tiny}             & -   & $\surd$ \\
~\cite{vedula2019towards}          & $\surd$ & -       & ~\cite{belouadah2019il2m}            & -   & $\surd$ \\
~\cite{vedula2020automatic}        & $\surd$ & -       & ~\cite{caccia2020online}             & -   & $\surd$ \\
~\cite{madaan2021representational} & -       & $\surd$ & ~\cite{ebrahimi2020remembering}      & -   & $\surd$ \\
~\cite{purushwalkam2022challenges} & -       & $\surd$ & ~\cite{saha2020gradient}             & -   & $\surd$ \\
~\cite{fini2022self}               & -       & $\surd$ & ~\cite{tang2021layerwise}            & -   & $\surd$ \\
~\cite{cha2021co2l}                & -       & $\surd$ & ~\cite{benzing2022unifying}          & -   & $\surd$ \\
~\cite{lopez2017gradient_gem}      & -       & $\surd$ & ~\cite{gong2022continual}            & -   & $\surd$ \\
\textit{\textbf{OpenCML}}                            & $\surd$ & $\surd$ &                                      &     &         \\ \hline
\end{tabular}
Abbreviation: OTC: Open-text classification  CML: Continual Machine Learning
\end{table}

\section{Proposed Methodology}
To answer $RQ1$, this section presents the functionality and provides a detailed description of OpenCML. The functionality of OpenCML can be broken down into four stages, each with its own set of tasks and functions. The first stage uses BERT (Bidirectional Encoder Representations from Transformers) to preprocess the input data. The second stage identifies unknown instances, which is based on CNN (Section~\ref{otc}). It discovers novel instances from test data and stores them in a separate memory. The third stage creates clusters from unknown instances discovered by stage 2, using the balanced iterative reducing and clustering using hierarchies (BIRCH) algorithm. Next, it creates novel classes using the key extraction method and labels these classes of unknown test data instances (Section~\ref{dinc}). Finally, it learns novel classes incrementally from the second iteration onward. OpenCML utilises cross-distillation loss to prevent forgetting and learn classes incrementally (Section~\ref{cml}).

\begin{figure*}
    \centering
    \includegraphics[height=4.0in, width =5.0in]{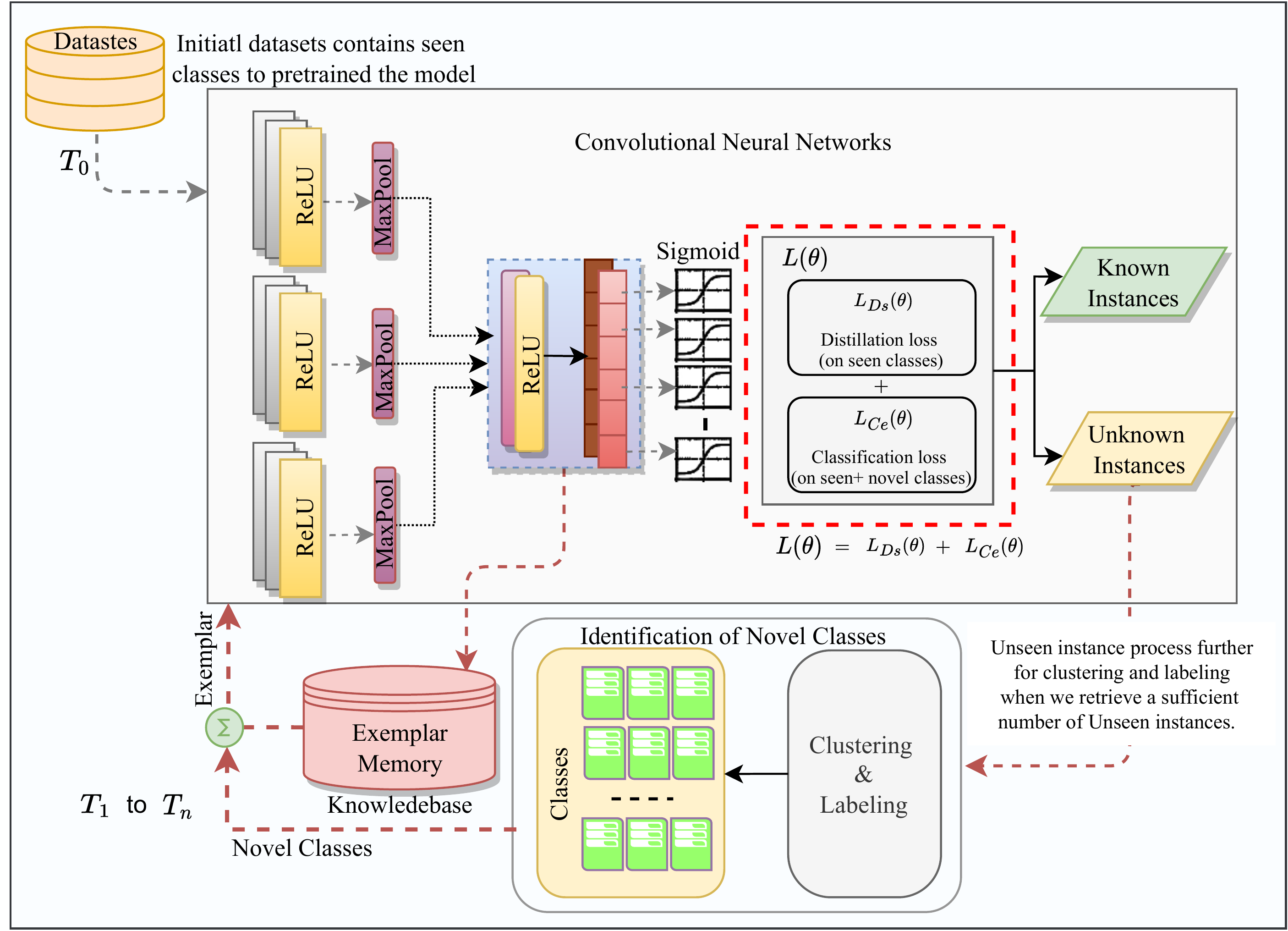}
    \caption{ This figure explains the functioning of the \textit{OpenCML} Framework. The Custom Loss $L(\theta)$ formulation, used to retain previously learned knowledge, is explained in a red dotted line box. Where $L_{Ds}(\theta)$  is distillation loss applied on old classification layer (on known classes), and $L_{Ce}(\theta)$  is classification loss which is applied on both new and old classification layer (on known and novel classes) } 
    \label{model}
\end{figure*}

\subsection{Open Text Classification} \label{otc}
Module one classifies unknown and known data using convolutional neural networks, which consist of three essential components for data classification. The first component of the data processing unit uses embedding for data embedding. Next, it consists of a convolutional network with max-pooling; the third component is the output unit. It is a fully connected layer. This module comprises convolution layers with a middle Rectified Linear Unit (ReLU) function to normalise the convolutional output. Next, the Max-pooling layer is used to decrease the dimensions of the convolution layer's output. Next, it consists of two fully interconnected layers with ReLU functions in between. The output layer uses a 1-vs-rest sigmoid layer. The description in detail is provided below.

Let $d_{im}$ is world vector (dimensional) $s_i \in \mathbb{R}^d_{im} $ corresponding to the $i^{th}$ word. Now let us assume length of sentence with padding is $l_p$. Now, the concatenation for every word vector is as follows:
\begin{equation}
    s_{i:l_p} = s_1 \mathbin\Vert s_2 \mathbin\Vert s_3 \mathbin\Vert, \ldots, s_{l_p}    
\end{equation}

Concatenation is denote by "$\mathbin\Vert$", therefore the general representation of "$\mathbin\Vert$" well be  $s_{i:i+j}$ that is concatenations of $s_i,\ s_i+1, \ldots s_{i+j}$. Now we apply the convolution filters $c \in \mathbb{R}^{w{d_{im}}}$ on word window with size $w$ and generates the feature map $F_m$ from the $s_{i:1+w-1}$ word window, that is represented as:
\begin{equation}
    {F_m}_i = f(c.s_{i:1+w-1}+b)
\end{equation}

Where the non-leaner features represented as $f$ and bias as $b$, that is $b \in \mathbb{R}$. we get the possible word window $s_{1:w}, s_{2:w},\ldots s_{{l_p}-w+1:l}$. After, applying filters $f$ on wold windows we get the feature map $F_m$, that is $F_{m} \in \mathbb{R}^{{l_p}-{w+1}}$.

\begin{equation}
  F_{m}= (F_1, F_2, \ldots, F_{{{l_p}-w+1}})                       
\end{equation}
After every convolution there is ReLU activation function to normalise the output of convolution layer. To reduce the $F_{m}$ (feature vector) it uses $1D$-max pool after convolution layer. It takes maximum value of $\hat{F_{m}}$  form convolution layer that is $ \hat{F_{m}}=Max(F_{m})$

To segregate the unknown instances from testing data, it uses the 1-vs-rest method. It can classify unknown samples using a 1-vs-rest layer. The conventional functions, such as softmax and other similar ones, are proven effective for multi-class classification, but they cannot reject unknown data instances.

Here, we used $S_{cn}$ (number of known classes ) sigmoid in the last layer. The $i^{th}$ sigmoid function for $\alpha _{i}$ classes.
Now model assigned classes as positive classes if $\beta = \alpha _{i}$ and classes denoted as negative for all $\beta  \neq  \alpha _{i}$.

\begin{equation}
    \hat{\beta} =\begin{Bmatrix}
reject, \  if sigmoid (\kappa _i )< \gamma_i ,\   \forall \ \alpha_i\ in \ \beta_i\\ 
argmax_{\alpha_i \in \beta} \  sigmoid(\kappa _i), \  Otherwise 
\end{Bmatrix}
\end{equation}

Where $I$ is the Indicator function, j= 1 to $I_c$ (Number of instances) and probability output of $S_{cn}$ sigmoid for $j^{th}$ input of $i^{th}$ dimension of  $\kappa$ is  $p(\beta_j = \alpha_i) = sigmoid(\kappa_{j,i})$

\subsection{Discovery and Identification of Novel Classes} \label{dinc}
To make classes from unknown data, we applied the Balanced Iterative Reducing and Clustering using Hierarchies (BIRCH) algorithm. We have $N_d$ data points with $d$ dimensions, each represented as a vector. Now, it calculates the cluster features $C_f$ for all $C_j={x_i \dots x_{N_{d}}}$. To calculate the cluster feature, $C_f$ consists of three parameters: $N_d$, $L_s$, and $S_s$. Where $N_d$ is the number of data points with d-dimensions,  $L_s$ is the linear sum, and $S_s$ is the squared sum. The $C_f$, $L_s$, and  $S_s$ can be calculated as: 

\begin{equation}
    C_f= \left ( N_d,\vec{L_s}, S_s \right)
\end{equation}

\begin{equation}
    \vec{L_s} =\sum_{i=1}^{N_d} \vec{x_i}
\end{equation}

\begin{equation}
    S_s = \sum_{i=1}^{N_d}(\vec{x_i})^2
\end{equation}

The cluster features are systematised in the tree, called a feature tree; it is a height-balanced tree. It has two parameters  $\beta_f$ (branching factor) and $T_h$ (threshold).

Every non-lead node holds at most $\beta_f$,  from $[C_{f_i}, Ch_i]$, where $Ch_i$ is a child node, and it is a pointer to its $i^{th}$ child node and $C_{f_i}$ defines the associated sub-cluster. A leaf node contains at most  $L$ accesses individually of the form $C_{f_i}$. Two pointers, previous and next, are used to chain all leaf nodes. The height of the tree relies on the threshold $T_{h}$. 
Next, the algorithm reviews all the leaf accesses in the initial   $C_f$ tree to reconstruct a smaller $C_f$  tree while extracting outliers and setting dense sub-clusters into bigger ones.  
For all leaf entries, an agglomerative hierarchical clustering algorithm is used directly to define sub-clusters based on their $C_f$ vectors.

To avoid minor and localised inaccuracies, we redistribute the data points to their closest seeds (calculate the centroid), which provides a new set of clusters. The data points which are distant from the seeds are considered outliers. The centroid can be computed as:

\begin{equation}
   Centroid (C)= \frac{\sum_{i=1}^{N_d}\vec{x_i}}{N_d} =\frac{\vec{L_s}}{N_d}
\end{equation}

\subsection{Continual Machine Learning} \label{cml}
To continually acquire knowledge, OpenCML employed the cross-distillation loss method~\cite{castro2018end}. OpenCML utilises exemplar memory to store samples from previously learned classes (section~\ref{EM}). For the subsequent output, it constructs training data (section~\ref{ctd}). In the next iteration, when new classes are added to the input data, it applies cross-distillation loss, distillation loss on the classification layer (for old classes), and multi-class cross-entropy loss for all classification layers (section~\ref{cdl}). Next, it again updates the exemplar memory (section~\ref{uxm}).

\subsubsection{Exemplar Memory} \label{EM}

To answer $RQ2$, the proposed approach utilises techniques that efficiently manage memory for storing and managing previous knowledge. We tackle the challenge of class incremental learning by first classifying both known and unknown instances from the data and creating novel classes from the unknown data. To further enhance learning efficiency, we selectively store a subset of the most representative examples from known classes for future use. We employ a memory module with a restricted capacity of $K$ examples to achieve this, ensuring we are conscious of the memory limitations. As more class instances are accumulated, the number of examples per class decreases. The number of examples per class, denoted by $n$, is determined by $n = [K/c]$, where c represents the number of classes stored in memory, and K represents memory capacity. This careful selection process of memory enables us to retain diverse examples from known classes while still being mindful of memory limitations. This makes our approach efficient and effective for incremental learning tasks in classes. We employed the herding technique for sample selection in our approach. 

\textbf{Herding}~\cite{welling2009herding}
is a powerful method for selecting exemplars in class incremental learning tasks. This technique involves iteratively selecting the sample closest to the current centroid of the selected samples. In other words, the aim is to find a set of K exemplars $Y = {y_1, \dots, y_K}$ that maximise the sum of cosine similarity between the exemplars and the mean vector of the selected samples. This similarity score is computed using the cosine similarity function $cos(y_j, x_i)$, where $y_j$ is an exemplar and $x_i$ is a sample. The mean vector of the chosen samples up to the $i^{th}$ iteration is computed as $\frac{1}{i} \sum_{k=1}^i x_k$. Herding is highly efficient and effective for selecting representative exemplars in class incremental learning tasks.

\subsubsection{Construction of the Training Data} \label{ctd}
The Input for the second iteration onward was created by integrating the  Exemplar memory and new data (Input). When new Input is shown, the system integrates the stored exemplars with the new Input to generate a representation, which is then compared to the stored exemplars to identify the closest match. This process enables the accurate classification of new instances, as the variability and context-specific features are essential. By integrating exemplar memory and new input data, the system can identify new instances of a class based on their resemblance to prior experiences, resulting in greater flexibility and accuracy in classification.

\subsubsection{Custom Loss}\label{cdl}

To answer $RQ3$ and $RQ-4$, we used the custom loss~\cite{castro2018end}, which was created by combining the distillation loss~\cite{hinton2015distilling} and cross-entropy loss. The distillation loss retrieves the knowledge from prior learned classes, and the cross-entropy loss learns to classify the recent classes. We employ cross-entropy loss for all classification layers, whereas distillation loss is only applied to the classification layers of the older classes. This enables the model to adjust its decision boundaries and improve its performance. By integrating these two loss functions, our method offers a more effective and efficient solution for training deep neural networks on classification tasks. The custom loss function $L(\theta)$ can be defined as:

\begin{equation}L(\theta) = L_{Ce}(\theta) + \sum_{Cl=1}^{co} L_{{Ds}_{cl}}(\theta),
\label{eq:Custom_loss} \end{equation}

Where $L_{Ce}(\theta)$ is cross-entropy loss, which is applied on all the classes (old + new), $L_{{Ds}_{cl}}$ is the distillation loss function of classification layer $cl$, and $co$  number of classification layers for the old classes.
The $L_{Ce}(\theta)$ can be defined as:
\begin{equation}
    L_{Ce}(\theta) = -\frac{1}{I_c} \sum_{i=1}^{I_c} \sum_{j=1}^{C} g_{ij} \log s_{ij}
\end{equation}
Where $s_i$ is a score acquired by using a sigmoid function on the \textit{logits} of a classification layer, for instance, $i$, $g_i$ is the ground truth, $i$, and $I_c$ and $C$ denote the number of instances and classes, respectively.
The distillation loss $L_{Ds}(\theta)$ can be  defined as:

\begin{equation}
    L_{Ds}(\theta) = -\frac{1}{I_c} \sum_{i=1}^{I_c} \sum_{j=1}^{C} gdst_{ij} \log qsdst_{ij},
\label{eq:distillation_loss}
\end{equation}
where, $gdst_i$ and $sdst_i$ are revised interpretations of $g_i$ and $s_i$, respectively. They are acquired by increasing $g_i$ and $s_i$ to the exponent $1/T$, as illustrated in~\cite{hinton2015distilling}, where $T$ is the distillation parameter.

\subsubsection{Updating the Exemplar Memory}\label{uxm}
As memory is limited, we update the memory to incorporate samples from new classes after training has occurred. We performed this step after training, which concerns extracting examples from the end of the sample set of each class. Since the examples are stored in a sorted list, this process takes little effort and can be performed easily. It should be noted that the samples that are removed during this step are never used again. This approach can optimise memory usage in the proposed framework and encourage the inclusion of novel classes in the model.

\subsection{Use case Scenario}

\begin{figure}
    \centering
    \includegraphics[height=2.5in, width =3.8in]{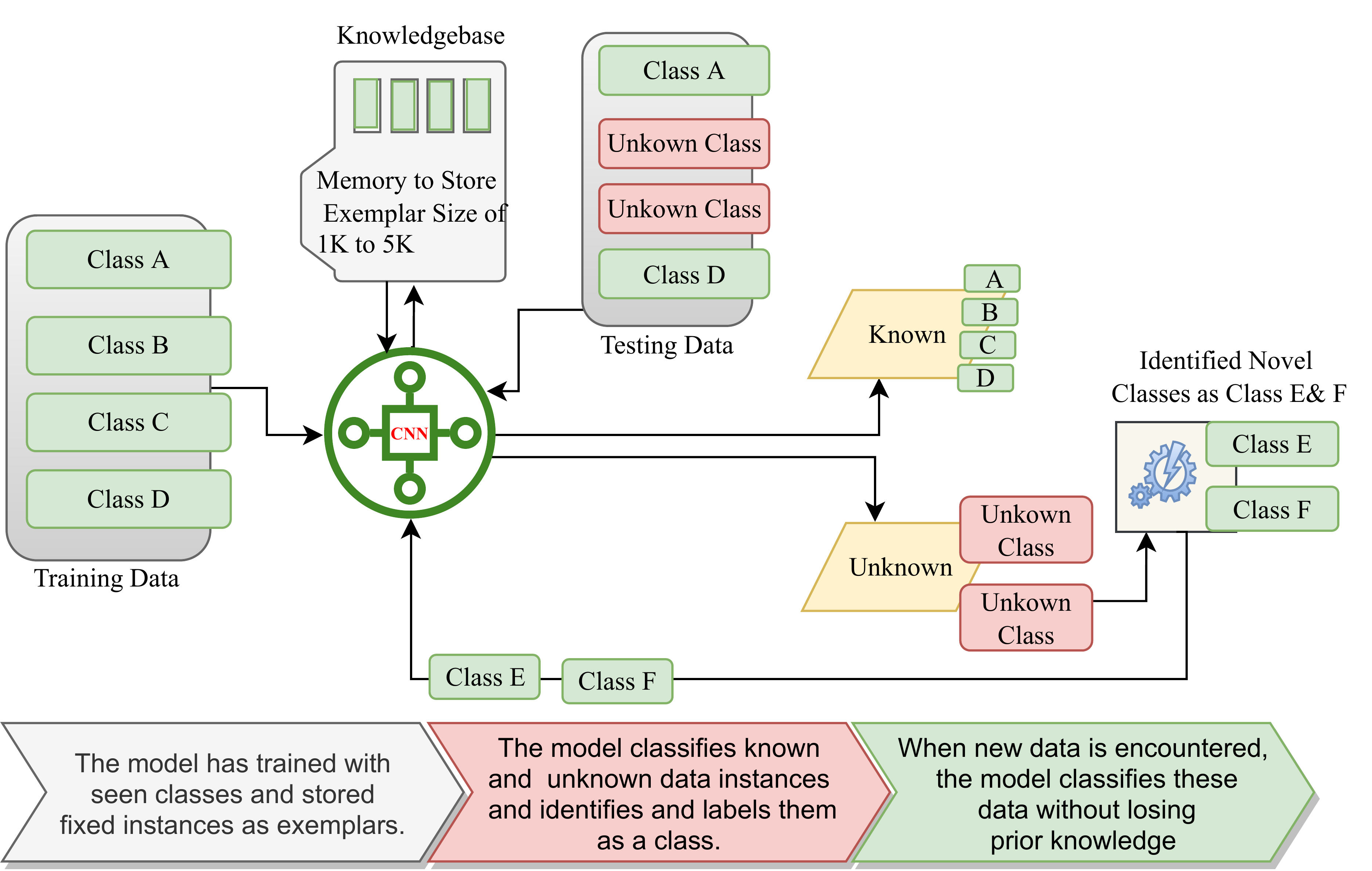}
    \caption{Use Case Scenario}
    \label{model1.2}
\end{figure}

To illustrate the methodology used in our research, let us consider an example of four distinct classes. However, only the classes $ A$, $ B$, $ C$, and $D$ were initially known, while $ E$ and $ F$ remained unidentified. Initially, our model performed class segregation by distinguishing between known and unknown classes. This segregation allowed us to classify the classes $A, B, C$ and $D$ as known, while E and F were classified as unknown.
Furthermore, we implemented a memory mechanism to store a fixed-size set of exemplars in the future. This memory served as a repository to retain valuable information from the initial process, ensuring its availability for subsequent classifications.
As our model progressed, it could identify additional novel classes, specifically classes $E$ and $F$, which were previously unknown. 
Subsequently, our model incorporated new data alongside the exemplars stored in memory. Additionally, it considered all six classes, $A, B, C, D, E$ and $F$, to refine its learning and classification capabilities. By incorporating these newly discovered classes, our model evolved and adapted to perform the overall classification task more effectively.
This iterative process, involving the integration of new data, the utilisation of exemplars from memory, and the continuous refinement of class identification, forms a crucial aspect of our research methodology. It enables the model to incrementally enhance performance and expand its understanding of the underlying data.

\section{Experiments and Results}
\subsection{ Experimental setup}
The proposed work's experiments were conducted on a computer equipped with an Intel Core i5-2410 M CPU and 8 GB of DDR3 RAM. The computer operated on a 64-bit version of Windows 10 with a 64-bit processor. All the experiments were implemented using Python 3.0.
To classify known and unknown classes, the threshold value $t_h =0.5$ is used for open text classification. We use Adam as the optimiser, and the weight decay for all parameters is set to 0.01. Each training step consists of 50 epochs with a learning rate of 0.02. Due to computational limitations, we have fixed four memory sizes, ranging from 250 to 1500, for storing examples. We conducted all experiments with up to four iterations.

\subsection{Datasets}
The evaluation was conducted using the text datasets that are publicly accessible, as listed below.
\begin{itemize}
     \item \textbf{BANKING77 (DS-1)}\footnote{https://github.com/PolyAI-LDN/task-specific-datasets/tree/master/banking\_data}: A dataset discussed by, is explicitly designed with banking sector domains. It covers a wide range of 77 distinct categories, serving as a valuable resource for classification tasks. The dataset is divided into three segments: the training dataset, encompassing 10,003 utterances; the validation dataset, containing 1,000 utterances; and the test dataset, which holds 3,080 utterances~\cite{casanueva2020efficient}.
     \item \textbf{CLINC150 (DS-2)}\footnote{https://github.com/clinc/oos-eval} : A dataset is specifically created for out-of-domain (OOD) detection tasks. It encompasses a diverse collection of 150 distinct classes drawn from 10 different domains. Within this dataset, there are 22,500 in-domain (IND) utterances representing the primary domain and an additional 1,200 out-of-domain (OOD) utterances, which are instances intended to challenge and evaluate the model's ability to detect OOD samples.~\cite{larson-etal-2019-evaluation}.

      \item \textbf{StackOverflow (DS-3)}: A dataset, originally introduced by~\cite{xu2015short}, this dataset is designed around 20 distinct categories and is thoughtfully segregated into three subsets to facilitate comprehensive model training and assessment. The training dataset comprises a substantial 12,000 data points, while the validation dataset, consisting of 2,000 data points, serves as a means for fine-tuning the model. Lastly, the test dataset includes 6,000 data points.

    \item\textbf{DBPedia Classes Dataset (DS-4)}\footnote{https://www.kaggle.com/danofer/dbpedia-classes} (DS-4):  This dataset includes more than 300k hierarchically labeled Wikipedia reports. It has three groups with 9, 70, and 219 classes~\cite{auer2007dbpedia}.
\end{itemize}

\subsection{Performance metric}
To evaluate the discovery of unknown classes, we employed several performance metrics, including Accuracy (\textit{Acc}), F1-score, Matthews Correlation Coefficient (M\textit{CC}), G-mean1 (\textit{GM1}), and G-mean2 (\textit{GM2}). Subsequently, to assess the effectiveness of labeling and keyword extraction for novel classes, we utilized metrics such as Accuracy (\textit{Acc}), Adjusted Rand Score (\textit{ARS}), Normalized Mutual Information (\textit{NMI}), Fowlkes-Mallows Score (\textit{FMS}), and \textit{F1-score}.

 Assuming standard terminology, True Positive ($T_{p_{os}}$), False Positive ($F_{p_{os}}$), True Negative ($T_{n_{eg}}$),   False Negative ($F_{n_{eg}}$), Precision ($P_{re}$), Recall ($R_{e}$), \& Specificity ($S_{pe}$). The formulas of performance matrices are as follows. 
 
\begin{table}[htb]
\centering
\scriptsize
\caption{Performance metrics }
\label{tab:PM}
\begin{tabular}{l|l}
\hline
\textbf{Parameter} & \textbf{Formula} \\ \hline
Accuracy  &= $\frac{T_{p_{os}}+T_{n_{eg}}}{T_{p_{os}}+T_{n_{eg}}+F_{p_{os}}+F_{n_{eg}}}$ \\ 
F1-score &= $2 \times \frac{P_{re}\times  R_{e}}{P_{re} +  R_{e}}$ \\ 
MCC &= $\frac{(T_{p_{os}} \times T_{n_{eg}})- (F_{p_{os}} \times F_{n_{eg}})}{\sqrt{(T_{p_{os}}+ F_{p_{os}})(T_{p_{os}} + f_{n_{eg}})(T_{n_{eg}} + F_{p_{os}})(T_{n_{eg}} + F_{n_{eg}})}}$ \\ 
G-mean 1 (GM1) &= $\sqrt{P_{re} \times R_{e}}$ \\ 
G-mean 2 (GM2) &= $\sqrt{R_{e} \times S_{pe}}$ \\ 
ARS &= $\frac{(RI - Expected RI)}{(max(RI) - Expected RI)}$ \\ 
NMI  &= $NMI(g_i ,g_{j}'')$ = $\sum_{i=1}^{|g_i|} \sum_{j=1}^{|g_{j}''|} \frac{|g _i\cap g {}''_j|}{N} \log\frac{N|g _i \cap g {}''_j|}{|g _i||g {}''_j|}$ \\ 
FMS  & = $\frac{{T_{p_{os}}}}{\sqrt {(T_{p_{os}}+F_{p_{os}})*(T_{p_{os}}+F_{n_{eg}})}}$ \\ \hline
\end{tabular}
\end{table}

\subsection{Performance Analysis}
The performance analysis conducted in this study encompasses three main aspects. Firstly, we examine the performance of class increment learning, which involves evaluating the accuracy across multiple rounds for each dataset. This analysis provides insights into the model's ability to learn new classes and adapt to evolving data incrementally. The results show variations in performance across the datasets. Second, we analyse the model's performance in the context of open-text classification during class incremental learning. This evaluation provides valuable insights into how the model performs open text classification tasks, particularly in the context of incremental learning. Considering these two aspects, we gain a comprehensive understanding of the model's overall performance and capabilities in class increment learning and open-text classification. Next, we conduct an ablation study.

\subsubsection{Class incremental learning}
The performance was measured using accuracy metrics: Accuracy of each iteration and Average accuracy. The experiment results are shown in Table~\ref{results}.

The analysis of incremental accuracy with varying memory bucket sizes (k) reveals significant insights into the model's performance across different iterations (I) of incremental learning. Specifically, the data shows that as the memory bucket size increases from k=250 to k=1500, the average incremental accuracy increases consistently. For DS-1, at k=250, the model acquires an average accuracy of 65.87\%, which increases to 68.34\% at k=500, 70.49\% at k=1000, and 71.99\% at k=1500. Notably, the incremental accuracy at the initial iteration (I-1, 14 classes) remains constant at 89.589 across all memory bucket sizes, indicating that the initial learning performance is unaffected by memory size. However, in succeeding iterations (I-2 to I-4), the model's accuracy declines as the number of classes increases, with a more prominent drop observed at smaller memory bucket sizes. For example, at k=250, the accuracy drops from 71.185 (I-2) to 42.453 (I-4), whereas at k=1500, it drops from 76.485 (I-2) to 54.748 (I-4). 

For DS-2, there are four incremental Iterations, I-1 to I-4, corresponding to an increasing number of classes (4, 6, 8, and 1-1 classes, respectively). 
For all memory bucket sizes, the model maintains a consistent accuracy of 91.898\% at the initial stage (I-1, four classes), indicating that the initial learning performance is independent of the memory size. With k=250, the accuracy drops from 80.785\% (I-2) to 60.254\% (I-4). In contrast, at k=1500, the accuracy decreases from 84.999\% (I-2) to 65.789
The average incremental accuracy shows a positive correlation with the size of the memory bucket. Specifically, the average accuracy improves from 75.847\% (k=250) to 77.104\% (k=500), 78.308\% (k=1000), and 79.606\% (k=1500). 

The DS-3 is evaluated across four incremental iterations, I-1 to I-4, corresponding to an increasing number of classes (3, 5, 7, and 9 classes, respectively). The memory bucket sizes under consideration are k=250, k=500, k=1000, and k=1500.
For each memory bucket size, the model's incremental accuracy at the initial stage (I-1,  three classes) remains constant at 89.987\%, indicating that the initial learning performance is not affected by the memory size, with k=250, the accuracy drops from 82.754\% (I-2) to 66.742\% (I-4), whereas with k=100, the accuracy decreases from 89.168\% (I-2) to 71.265\% (I-4).
The average incremental accuracy shows a positive correlation with the size of the memory bucket. Specifically, the average accuracy increases from 77.985\% (k=250) to 79.217\% (k=500), 80.178\% (k=1000), and 82.541\% (k=1500). 

The DS-4 is evaluated over four incremental Iterations, I-1 to I-4, which correspond to an increasing number of classes (5, 7, 9, and 11 classes, respectively). The memory bucket sizes considered are k=250, k=500, k=1000, and k=1500. At the initial stage (I-1, five classes), the model consistently executes an incremental accuracy of 90.956\% across all memory bucket sizes, indicating that the initial performance is unaffected by the size of the memory bucket, with k=250, the accuracy drops from 72.58\% (I-2) to 51.316\% (I-4), whereas with k=1500, the accuracy decreases from 81.889\% (I-2) to 59.124\% (I-4).
The average incremental accuracy demonstrates a positive correlation with the size of the memory bucket. Specifically, the average accuracy improves from 68.828\% (k=250) to 71.593\% (k=500), 73.019\% (k=1000), and 74.810\% (k=1500).To the best of our knowledge, there is no work available that contains both open-text classification and continual learning. Therefore, we present a summarized evaluation in Table \ref{theoryCom}. Table~\ref{theoryCom} clearly states that only a few existing works are available in this domain for natural language processing (NLP) in continual machine learning. Moreover, based on our extensive review and knowledge in the field, we have yet to find research that comprehensively addresses the combined challenges of open-text classification and continual machine learning. This research gap underscores the need for novel approaches to tackle these challenges.

\begin{table}[]
\caption{Performance analysis of OpenCML with four benchmark datasets. It shows the average results of four rounds (Iterations, that is indicated by I, $I_1$ to $I_3$) with M@250 to M@1500, where M= exemplar memory}
\label{results}
\centering
\begin{tabular}{lccccl} \\ \hline
\multicolumn{6}{c}{DS-1}                                                                                                                                                                                                                                                                                                                                                                                                                                                                                              \\ 
\begin{tabular}[c]{@{}l@{}}Memory \\ Bucket\end{tabular}                          & \begin{tabular}[c]{@{}l@{}}I-1\\ (14-classes)\end{tabular}                         & \begin{tabular}[c]{@{}l@{}}I-2\\ (18-classes)\end{tabular}                         & \begin{tabular}[c]{@{}l@{}}I-3\\ (19-classes)\end{tabular}                         & \begin{tabular}[c]{@{}l@{}}I-4\\ (20-classes)\end{tabular}                          & \begin{tabular}[c]{@{}l@{}}Avg. Incremental\\ Accuracy\end{tabular}                          \\ \hline 
k=250                                                                             & 89.589                                                                            & 71.185                                                                            & 60.248                                                                            & 42.4527                                                                            & 65.868                                                                                    \\
k=500                                                                             & 89.589                                                                            & 73.145                                                                            & 61.164                                                                            & 49.475                                                                             & 68.343                                                                                     \\
k=1000                                                                            & 89.589                                                                            & 75.963                                                                            & 64.968                                                                            & 51.427                                                                             & 70.488                                                                                     \\
k=1500                                                                            & 89.589                                                                            & 76.485                                                                            & 67.154                                                                            & 54.748                                                                             & 71.994                                                                                       \\ \hline 
\multicolumn{6}{c}{DS-2}                                                                                                                                                                                                                                                                                                                                                                                                                                                                                              \\
\begin{tabular}[c]{@{}l@{}}Memory \\ Bucket\end{tabular}                          & \begin{tabular}[c]{@{}l@{}}I-1\\ (4-classes)\end{tabular}                          & \begin{tabular}[c]{@{}l@{}}I-2\\ (6-classes)\end{tabular}                          & \begin{tabular}[c]{@{}l@{}}I-3\\ (8-classes)\end{tabular}                          & \begin{tabular}[c]{@{}l@{}}I-4\\ (11-classes)\end{tabular}                          & \begin{tabular}[c]{@{}l@{}}Avg. Incremental\\ Accuracy\end{tabular}                          \\ \hline 
k=250                                                                             & 91.898                                                                            & 80.785                                                                            & 70.452                                                                            & 60.254                                                                             & 75.847                                                                                     \\
k=500                                                                             & 91.898                                                                            & 81.988                                                                            & 72.784                                                                            & 61.745                                                                             & 77.103                                                                                     \\
k=1000                                                                            & 91.898                                                                            & 83.745                                                                            & 73.965                                                                            & 63.625                                                                             & 78.308                                                                                     \\
k=1500                                                                            & 91.898                                                                            & 84.999                                                                            & 75.737                                                                            & 65.789                                                                             & 79.606                                                                                     \\ \hline 
\multicolumn{6}{c}{DS-3}                                                                                                                                                                                                                                                                                                                                                                                                                                                                                         \\
\begin{tabular}[c]{@{}l@{}}Memory \\ Bucket\end{tabular}                          & \begin{tabular}[c]{@{}l@{}}I-1\\ (3-classes)\end{tabular}                          & \begin{tabular}[c]{@{}l@{}}I-2\\ (5-classes)\end{tabular}                          & \begin{tabular}[c]{@{}l@{}}I-3\\ (7-classes)\end{tabular}                          & \begin{tabular}[c]{@{}l@{}}I-4\\ (9-classes)\end{tabular}                           & \begin{tabular}[c]{@{}l@{}}Avg. Incremental\\ Accuracy\end{tabular}                          \\ \hline 
k=250                                                                              & 89.987                                                                            & 82.754                                                                            & 72.457                                                                            & 66.742                                                                             & 77.985                                                                                       \\
k=500                                                                              & 89.987                                                                            & 83.412                                                                            & 75.634                                                                            & 67.835                                                                             & 79.217                                                                                       \\
k=1000                                                                              & 89.987                                                                            & 86.879                                                                            & 74.498                                                                            & 69.348                                                                             & 80.178                                                                                       \\
k=1500                                                                             & 89.987                                                                            & 89.168                                                                            & 79.743                                                                            & 71.265                                                                             & 82.541                                                                                     \\ \hline 
\multicolumn{6}{c}{DS-4}                                                                                                                                                                                                                                                                                                                                                                                                                                                                                       \\ 
\multicolumn{1}{c}{\begin{tabular}[c]{@{}c@{}}Memory \\      Bucket\end{tabular}} & \multicolumn{1}{c}{\begin{tabular}[c]{@{}c@{}}I-1\\      (5-classes)\end{tabular}} & \multicolumn{1}{c}{\begin{tabular}[c]{@{}c@{}}I-2\\      (7-classes)\end{tabular}} & \multicolumn{1}{c}{\begin{tabular}[c]{@{}c@{}}I-3\\      (9-classes)\end{tabular}} & \multicolumn{1}{c}{\begin{tabular}[c]{@{}c@{}}I-4\\      (11-classes)\end{tabular}} & \multicolumn{1}{c}{\begin{tabular}[c]{@{}c@{}}Avg. Incremental\\      Accuracy\end{tabular}} \\ \hline 
\multicolumn{1}{c}{k=250}                                                         & \multicolumn{1}{c}{90.956}                                                        & \multicolumn{1}{c}{72.58}                                                         & \multicolumn{1}{c}{60.458}                                                        & \multicolumn{1}{c}{51.316}                                                         & \multicolumn{1}{c}{68.827}                                                                  \\
\multicolumn{1}{c}{k=500}                                                         & \multicolumn{1}{c}{90.956}                                                        & \multicolumn{1}{c}{77.826}                                                        & \multicolumn{1}{c}{62.625}                                                        & \multicolumn{1}{c}{54.963}                                                         & \multicolumn{1}{c}{71.592}                                                                  \\
\multicolumn{1}{c}{k=1000}                                                        & \multicolumn{1}{c}{90.956}                                                        & \multicolumn{1}{c}{79.245}                                                        & \multicolumn{1}{c}{65.731}                                                        & \multicolumn{1}{c}{56.145}                                                         & \multicolumn{1}{c}{73.019}                                                                 \\
\multicolumn{1}{c}{k=1500}                                                        & \multicolumn{1}{c}{90.956}                                                        & \multicolumn{1}{c}{81.889}                                                        & \multicolumn{1}{c}{67.269}                                                        & \multicolumn{1}{c}{59.124}                                                         & \multicolumn{1}{c}{74.809}       \\ \hline                                                           
\end{tabular}
\end{table}

\begin{landscape}
\begin{table}[htb]
\scriptsize
\caption{Incremental Open Classification Performance Analysis of Proposed Model}
\label{tab:OOA}
\begin{tabular}{p{1.5cm}p{1.0cm}p{1.0cm}p{1.0cm}p{1.0cm}p{1.0cm}p{1.0cm}p{1.0cm}p{1.0cm}p{1.0cm}p{1.0cm}p{1.0cm}p{1.0cm}p{1.0cm}} \hline
     &         & Banking77 &        &         & CLINC 150 &        &         & StackOverFlow &        &         & DBPedia Classes &        &         \\ \hline
     &         & ACC-ALL   & F1-OOD & F1-KNOW & ACC-ALL   & F1-OOD & F1-KNOW & ACC-ALL       & F1-OOD & F1-KNOW & ACC-ALL         & F1-OOD & F1-KNOW \\ \hline
     
     & MSP     & 26.72     & 9.03   & 50.09   & 30.54     & 24.25  & 39.55   & 42.45         & 44.43  & 42.95   & 44.89           & 46.76  & 45.7    \\
     & DOC     & 29.86     & 14.92  & 45.62   & 45.56     & 48.81  & 47      & 26.93         & 8.89   & 44.16   & 29.34           & 6.95   & 40.89   \\
     & OpenMax & 81.69     & 87.11  & 72.72   & 89.79     & 93.42  & 79.69   & 91.53         & 94.41  & 83.18   & 88.28           & 91.43  & 85.92   \\
     & Softmax & 79.42     & 84.87  & 72.06   & 89.78     & 93.34  & 81.74   & 89.76         & 93.6   & 73.62   & 92.74           & 90.32  & 75.89   \\
25\% & LMCL    & 29.06     & 13.83  & 50.66   & 32.38     & 27.45  & 41.05   & 58            & 64.81  & 51.8    & 61.36           & 61.05  & 53.97   \\
     & SEG     & 33.38     & 22.14  & 48.67   & 50.67     & 55.8   & 49.54   & 27.23         & 9.66   & 47.61   & 30.29           & 11.79  & 45.23   \\
     & ADB     & 85.32     & 89.8   & 78.06   & 90.61     & 93.95  & 81.36   & 93.2          & 95.58  & 85.3    & 96.17           & 98.24  & 88.79   \\
     & KNN     & 87.41     & 91.52  & 77.7    & 92.71     & 95.42  & 83.81   & 92.04         & 94.76  & 84.29   & 89.61           & 92.22  & 86.32   \\ \hline
I-1  & Ours    & 89.73     & 78.18  & 93.85   & 85.69     & 91.45  & 62.22   & 95.47         & 83.58  & 95.86   & 93.18           & 81.22  & 92.14   \\
I-2  & Ours    & 91.65     & 78.15  & 93.8    & 77.56     & 55.45  & 86.88   & 81.56         & 58.91  & 92.38   & 84.42           & 55.6   & 89.24   \\
I-3  & Ours    & 92.89     & 79.68  & 93.98   & 88.69     & 72.59  & 92.56   & 87.28         & 58.67  & 92      & 83.69           & 60.77  & 94.89   \\
I-4  & Ours    & 91.37     & 80.67  & 94.85   & 90.86     & 77.68  & 94.86   & 86.27         & 63.89  & 92.67   & 89.88           & 60.77  & 90.09   \\ \hline
     &         &           &        &         &           &        &         &               &        &         &                 &        &         \\
     & MSP     & 47.89     & 4.37   & 68.49   & 42.51     & 12.89  & 62.36   & 60            & 52.92  & 68.87   & 57.99           & 49.53  & 66.13   \\
     & DOC     & 49.98     & 11.82  & 68.26   & 55.44     & 43.8   & 67.19   & 47.92         & 7.93   & 65.49   & 50.5            & 5.39   & 68.51   \\
     & OpenMax & 80.9      & 81.32  & 81.79   & 88.61     & 90.62  & 86.52   & 88.52         & 89.57  & 87.13   & 90.74           & 92.75  & 83.97   \\
     & Softmax & 80.32     & 80.57  & 81.5    & 87.91     & 89.71  & 87.03   & 83.47         & 85.48  & 80.31   & 86.7            & 87.93  & 77.63   \\
50\% & LMCL    & 50.45     & 12.63  & 69.56   & 46.53     & 24.26  & 63.6    & 63.25         & 58.03  & 72.9    & 60.59           & 60.38  & 75.19   \\
     & SEG     & 50.61     & 12.33  & 70      & 58.19     & 49.04  & 68.72   & 49.13         & 12.11  & 67.56   & 46.02           & 15.75  & 63.85   \\
     & ADB     & 81.86     & 81.51  & 83.9    & 89.5      & 91.4   & 87.49   & 89.45         & 90.46  & 88.47   & 86.29           & 93.17  & 92.4    \\
     & KNN     & 81.98     & 81.65  & 84.03   & 89.96     & 91.72  & 88.15   & 88.92         & 89.69  & 88.09   & 92.46           & 86.59  & 90.95   \\ \hline
I-1  & Ours    & 84.98     & 77.92  & 89.57   & 88.78     & 82.98  & 91.78   & 92.67         & 88.3   & 94.89   & 96.3            & 90.6   & 98.2    \\
I-2  & Ours    & 87.78     & 84.4   & 93.59   & 82.11     & 76     & 85.8    & 78.77         & 60.67  & 86.58   & 82.12           & 58.3   & 90.11   \\
I-3  & Ours    & 92.73     & 87.79  & 94.97   & 89.62     & 84.78  & 92.28   & 77.24         & 59.8   & 85.78   & 79.42           & 62.11  & 88.38   \\
I-4  & Ours    & 91.81     & 89.47  & 94.78   & 89.89     & 84.12  & 91.98   & 84.96         & 76.56  & 90.39   & 88.61           & 74.11  & 88.38   \\ \hline
     &         &           &        &         &           &        &         &               &        &         &                 &        &         \\
     & MSP     & 72.31     & 13.32  & 83.53   & 57.82     & 7.47   & 75.24   & 69.58         & 16.52  & 80.86   & 23.89           & 6.14   & 53.62   \\
     & DOC     & 71.48     & 7.97   & 82.92   & 69.54     & 47.04  & 80.29   & 68.81         & 5.39   & 80      & 26.84           & 16.99  & 42.11   \\
     & OpenMax & 82.79     & 71.95  & 87.2    & 87.7      & 85.86  & 89.33   & 83.75         & 75.21  & 87.66   & 85.51           & 84.34  & 70.39   \\
     & Softmax & 82.06     & 65.6   & 87.68   & 88.59     & 87.18  & 89.68   & 83.44         & 73.62  & 87.53   & 82.03           & 81.44  & 69.06   \\
75\% & LMCL    & 73.64     & 19.58  & 84.61   & 59.7      & 13.06  & 76.19   & 71.07         & 20.09  & 81.61   & 26.74           & 11.34  & 54.46   \\
     & SEG     & 72.23     & 10.97  & 83.68   & 71.6      & 52.2   & 81.51   & 69.51         & 8.24   & 81.23   & 36.64           & 19.96  & 44.78   \\
     & ADB     & 83.3      & 69.03  & 88.09   & 88.59     & 86.45  & 90.6    & 84.53         & 73.75  & 88.35   & 88.03           & 93.16  & 80.51   \\
     & KNN     & 84.47     & 72.64  & 88.66   & 89.88     & 88.21  & 91.41   & 85            & 75.76  & 88.71   & 91.01           & 89     & 74.08   \\ \hline
I-1  & Ours    & 81.58     & 83.9   & 87.48   & 85.29     & 80.77  & 86.79   & 93.78         & 94.78  & 94.6    & 93.4            & 74.52  & 91.32   \\
I-2  & Ours    & 88.85     & 85.7   & 90.67   & 84.33     & 81.2   & 85.57   & 70.87         & 59.79  & 79.87   & 95.14           & 81.92  & 97.69   \\
I-3  & Ours    & 92.25     & 91.47  & 93.87   & 86.57     & 84.9   & 84.2    & 71.58         & 59.85  & 80.77   & 96.57           & 77.66  & 96.84   \\
I-4  & Ours    & 93.38     & 92.28  & 96.91   & 88.65     & 86.25  & 89.44   & 61.84         & 69.58  & 83.37   & 94.45           & 83.55  & 98.31  \\ \hline
\end{tabular}
\end{table}
\end{landscape}

\subsubsection{Open-text classification}
In our study, we conducted experiments with our model at three different levels of openness: 25\%, 50\%, and 75\%. Our preliminary research assessed our model's performance in open-text classification scenarios. To provide a comprehensive evaluation and benchmark the effectiveness of our model, we compared it with several methods widely used in open-text classification, such as; MSP~\cite{hendrycks2016baseline}, DOC~\cite{shu2017doc}, OpenMax~\cite{bendale2016towards},  Softmax~\cite{zhang2021deep},  LMCL~\cite{lin2019deep},  SEG~\cite{yan2020unknown},  ADB~\cite{zhang2021deep}, 
SCL~\cite{zeng2021modeling} SCL with Gaussian discriminant analysis (GDA) and SCL with Local Outlier Factor (LOF) these are proposed in KNN~\cite{zhou2022knn}.

The objective of comparing our model with these existing approaches is to understand how it performs compared to established techniques and to identify any potential improvements or advantages it may offer. The existing methods selected for comparison are recognised for their applicability in open-text classification tasks, serving as valuable reference points for evaluating the capabilities of our model.
These comparisons enabled us to gain meaningful insights into the strengths and weaknesses of our model in various scenarios of openness. By comparing our results with existing methods, we can evaluate whether our model's overall performance is improved despite introducing new data in each iteration. 
The choice of openness levels (25\%, 50\%, and 75\%) reflects the varying degrees of openness typically encountered in real-world text classification problems. This multilevel assessment allowed us to evaluate the adaptability and robustness of our model across a spectrum of open-text classification challenges, ranging from relatively constrained to highly open scenarios.

Table~\ref{tab:OOA} 
Show the incremental open classification performance analysis of the proposed model with Banking77, CLINC 150, StackOverFlow, and DBPedia classes.

\textit{Three Levels of Openness:} 25\%, 50\%, and 75\%: The analysis is carried out at different levels of openness. In this context, openness probably refers to the proportion of novel or previously unseen data introduced in each analysis. Initially, the openness is 25\%; that is, 25\% of the data introduced to the model is unknown and is not part of the training data. This increases to 50\% and 75\% openness in subsequent iterations.

\textit{Incremental Open Classification Performance Analysis:} This analysis involves gradually introducing more challenging open data into the model and observing how it performs as the level of openness increases. Open classification typically refers to a scenario in which the model needs to classify known or familiar classes (in-distribution data) and detect and handle previously unseen or novel classes (out-of-distribution data). This incremental approach helps assess the model's ability to adapt and generalize to new and unexpected information.

The finding is that the model's performance improves despite the inclusion of new data after every iteration. It indicates that the model is learning to adapt to new and unexpected data, a critical capability in many real-world applications where data are dynamic and ever-changing. The incremental open classification performance analysis demonstrates that the proposed model is becoming more robust and effective at handling new data and increasing openness to data. This ability to adapt and improve its performance despite encountering novel data is a positive sign of the model's versatility and generalisation capabilities.

\subsubsection{Ablation Study}
A study on the ablation of clustering and keyword extraction has been conducted to validate the used techniques. The detailed study and evaluation are available in the supplementary file\footnote{https://github.com/jitendraparmar94/OpenCml}.  

\subsubsection{performance analysis for clustering technique}
We evaluated the effectiveness of our clustering method by comparing it with the well-known k-means clustering algorithm. We compared the cluster sizes for K = 2, K = 3, and K = 4. Our evaluation metrics included Completeness, Homogeneity, and v-measure. Completeness measures the fraction of data points correctly assigned to the same cluster by the clustering algorithm and accurate labelling. Homogeneity measures the fraction of data points that belong to the same cluster in both the clustering result and precise labelling. V-measure is the harmonic mean of Completeness and Homogeneity, providing an overall measure of how well the clustering result matches the actual labelling.

With the CLINC-150 dataset, the analysis reveals that BIRCH outperforms K-means in terms of Homogeneity, Completeness, and V-measure for all cluster sizes. For K = 2, K-means has a higher homogeneity score of 0.773 and a lower completeness score of 0.636, while BIRCH has a lower homogeneity score of 0.859 and a slightly higher completeness score of 0.651. The V-measure score for K-means is 0.528, while the V-measure score for BIRCH is 0.562. For K=3, K-means has a higher homogeneity score of 0.810 and a lower completeness score of 0.491, while BIRCH has a slightly higher homogeneity score of 0.842 and a higher completeness score of 0.718. The V-measure score for K-means is 0.555, while the V-measure score for BIRCH is 0.596. For K = 4, K-means has a lower homogeneity score of 0.793 and a higher completeness score of 0.729, whereas BIRCH has a higher homogeneity score of 0.845 and a slightly higher completeness score of 0.751. The V-measure score for K-means is 0.614, while the V-measure score for BIRCH is 0.633.

 With the SNIPS dataset, for K = 2, the K-means algorithm achieves a homogeneity score of 0.666 and a completeness score of 0.630. In contrast, the BIRCH algorithm obtains a homogeneity score of 0.730 and a completeness score of 0.695. For the V-measure, the K-means algorithm yields a score of 0.422, whereas the BIRCH algorithm achieves a score of 0.520.
For K=3, the K-means algorithm has a homogeneity score of 0.491, completeness score of 0.512, and V-measure score of 0.554, while the BIRCH algorithm has a homogeneity score of 0.543, completeness score of 0.636, and V-measure score of 0.583.
For K=4, the K-means algorithm achieves a homogeneity score of 0.760, a completeness score of 0.788, and a V-measure score of 0.616. In contrast, the BIRCH algorithm obtains a homogeneity score of 0.789, a completeness score of 0.807, and a V-measure score of 0.668. Overall, BRICH outperforms.

With GOOGLE SNIPPETS. For K=2, K-means has a homogeneity score of 0.367 and a completeness score of 0.530, while BIRCH has a homogeneity score of 0.516 and a completeness score of 0.736. The V-measure score for K-means is 0.521, while the V-measure score for BIRCH is 0.642. For K=3, K-means has a higher homogeneity score of 0.591 and a completeness score of 0.651, while BIRCH has a lower homogeneity score of 0.658 and a higher completeness score of 0.765. The V-measure score for K-means is 0.694, while the V-measure score for BIRCH is 0.673. For K=4, K-means has a higher homogeneity score of 0.687 and a lower completeness score of 0.602, while BIRCH has a lower homogeneity score of 0.698 and a higher completeness score of 0.800. The V-measure score for K-means is 0.737, while the V-measure score for BIRCH is 0.891.

With DBPedia, we observed that for K = 2, K-means has a higher homogeneity score of 0.627 and a lower completeness score of 0.530. In contrast, BIRCH has a lower homogeneity score of 0.681 and a higher completeness score of 0.695. The V-measure score for K-means is 0.521, while the V-measure score for BIRCH is 0.620. For K = 3, K-means has a lower homogeneity score of 0.629 and a higher completeness score of 0.637. BIRCH has a slightly higher homogeneity score of 0.641 and a marginally higher completeness score of 0.664. The V-measure score for K-means is 0.678, while the V-measure score for BIRCH is 0.765. For K=4, K-means has a higher homogeneity score of 0.743 and a slightly lower completeness score of 0.754, while BIRCH has a lower homogeneity score of 0.784 and a higher completeness score of 0.805. The V-measure score for K-means is 0.789, while the V-measure score for BIRCH is 0.855.

Based on the clustering performance evaluation using the homogeneity, completeness, and V-measure metrics for different values of K (2, 3, and 4), it can be concluded that the BIRCH algorithm outperforms the K-means algorithm for the given experiment. The results demonstrate that BIRCH achieves higher homogeneity, completeness, and V-measure scores across all values of K.
Therefore, based on these findings, using the BIRCH clustering algorithm for this particular experiment is recommended. These results offer valuable insights into selecting the most suitable clustering algorithm for similar experiments in the future.

\begin{table}[htb]
\scriptsize
\caption{Performance analysis of clustering with K=2, K=3, and K=4. Where K is number of clusters.}
\label{undefined}
\begin{tabular}{lllllll}
\hline
\multicolumn{7}{c}{DS-1}                                                                 \\ \hline
             & \multicolumn{2}{c}{K=2}  & \multicolumn{2}{c}{K=3}  & \multicolumn{2}{c}{K=4}  \\ \hline
             & K-means & \textit{BIRCH} & K-means & \textit{BIRCH} & K-means & \textit{BIRCH} \\ \hline
Homogeneity  & 0.773   & 0.859          & 0.810   & 0.842          & 0.793   & 0.845          \\
Completeness & 0.636   & 0.651          & 0.491   & 0.718          & 0.729   & 0.751          \\
V-Measure    & 0.528   & 0.562          & 0.555   & 0.596          & 0.614   & 0.633          \\ \hline
\multicolumn{7}{c}{DS-2}                                                           \\ \hline
Homogeneity  & 0.666   & 0.730          & 0.491   & 0.543          & 0.760   & 0.789          \\
Completeness & 0.630   & 0.695          & 0.512   & 0.636          & 0.788   & 0.807          \\
V-Measure    & 0.422   & 0.520          & 0.554   & 0.583          & 0.616   & 0.668          \\ \hline
\multicolumn{7}{c}{DS-3}                                                 \\ \hline
Homogeneity  & 0.367   & 0.516          & 0.591   & 0.658          & 0.687   & 0.698          \\
Completeness & 0.530   & 0.736          & 0.651   & 0.765          & 0.602   & 0.800          \\
V-Measure    & 0.521   & 0.642          & 0.694   & 0.673          & 0.737   & 0.891          \\ \hline
\multicolumn{7}{c}{DS-4}                                                          \\ \hline
Homogeneity  & 0.627   & 0.681          & 0.629   & 0.641          & 0.743   & 0.784          \\
Completeness & 0.530   & 0.695          & 0.637   & 0.664          & 0.754   & 0.805          \\
V-Measure    & 0.521   & 0.620          & 0.678   & 0.765          & 0.789   & 0.855          \\ \hline
\end{tabular}
\end{table}

\subsubsection{performance analysis for different Labelling techniques}
The technique used to discover unknown examples and identify novel classes has employed BRICH and the single-rank method. However, we evaluated the OpenCML framework with different approaches to validate the performance. We conducted an ablation study to justify the use of particular techniques.

For keyword extraction, we employed the Single Rank method, which outperforms contemporary methods and KeyBERT, a recently proposed method. We have compared six different keyword extraction methods, as evaluated by several metrics, including Accuracy, Adjusted Rand Score (ARS), Normalised Mutual Information (NMI), Fowlkes-Mallows Score (FMS), and F1-score. 
For DS-1, it is clear from the graph that the SingleRank method outperforms all other methods in terms of Accuracy, NMI, FMS, and F1-score, with an accuracy score of 0.879 and an NMI score of 0.682. TextRank also performs well, with an accuracy score of 0.856 and an NMI score of 0.525. YAKE and KeyBERT demonstrate similar performance, with YAKE achieving higher scores in ARS, NMI, and FMS, while KeyBERT exhibits a higher accuracy score. Kea presents the lowest overall performance, with insufficient Accuracy, ARS, and F1-score scores. Overall, Singlerank performed better than other methods.

For DS-2, SingleRank achieves the highest F1 score of 0.989, the highest among all the algorithms. Additionally, SingleRank achieves the highest scores for Accuracy and NMI, at 0.979 and 0.917, respectively. SingleRank's score for ARS is slightly lower than YAKE, but it still achieves a high score of 0.952, indicating its effectiveness for automated keyword extraction. In comparison, other algorithms, such as TFIDF and Kea, have significantly lower scores across all metrics, indicating their lower effectiveness compared to SingleRank. It indicates that SingleRank is highly effective and outperforms other algorithms across all the evaluation metrics.

For DS-3, SingleRank achieves the highest score for ARS with a score of 0.682, the highest among all algorithms. SingleRank also achieves the highest F1-score, with a score of 0.888, surpassing YAKE, TextRank, Kea, and KeyBERT. Additionally, SingleRank achieves a higher score than TextRank and Kea in terms of Accuracy, with a score of 0.899. However, YAKE has the highest accuracy score, at 0.936. In contrast, TFIDF and Kea have the lowest scores across most metrics. The YAKE performance is slightly better here, but the Singlerank is given a better F1-score. 

For DS-4, Singlerank achieves the highest Accuracy, ARS, F1-score, and FMS scores, with scores of 0.742, 0.512, 0.712, and 0.678, respectively. While its score for NMI is the same as that of YAKE and Kea, it still outperforms TextRank and KeyBERT in this metric. In comparison, YAKE and Kea have the lowest scores for Accuracy, with scores of 0.562. Similarly, TextRank has the lowest score for ARS and the second-lowest score for Accuracy. KeyBERT also has a lower score for ARS than SingleRank. Overall, SingleRank is the most effective and outperforms other algorithms across most metrics.

\begin{figure*}[!htb]
    \centering
    \begin{subfigure}[b]{0.45\textwidth}
        \centering
        \includegraphics[width=\textwidth]{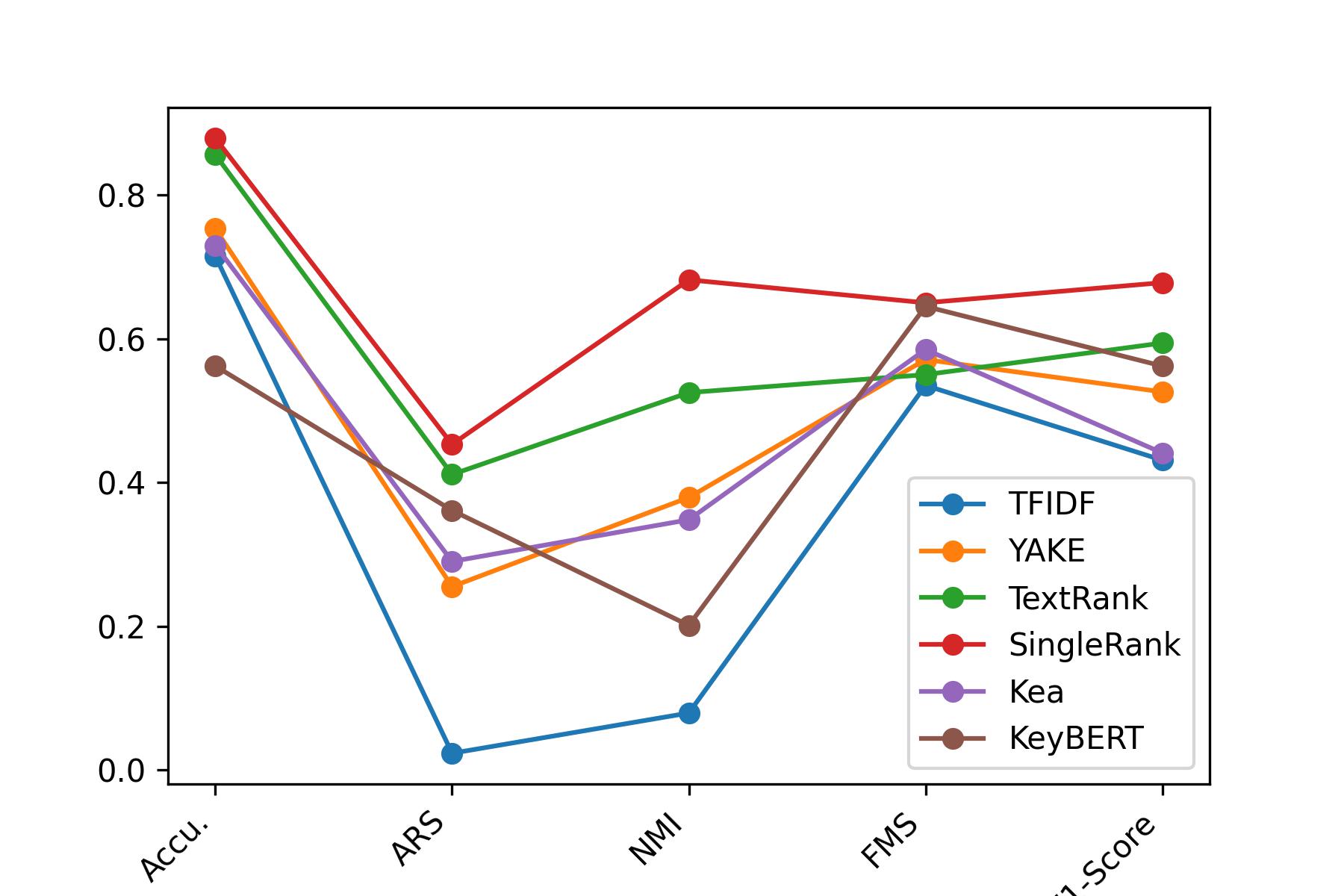}
        \caption{DS-1}
        \label{fig:img1}
    \end{subfigure}
    \hfill
    \begin{subfigure}[b]{0.45\textwidth}
        \centering
        \includegraphics[width=\textwidth]{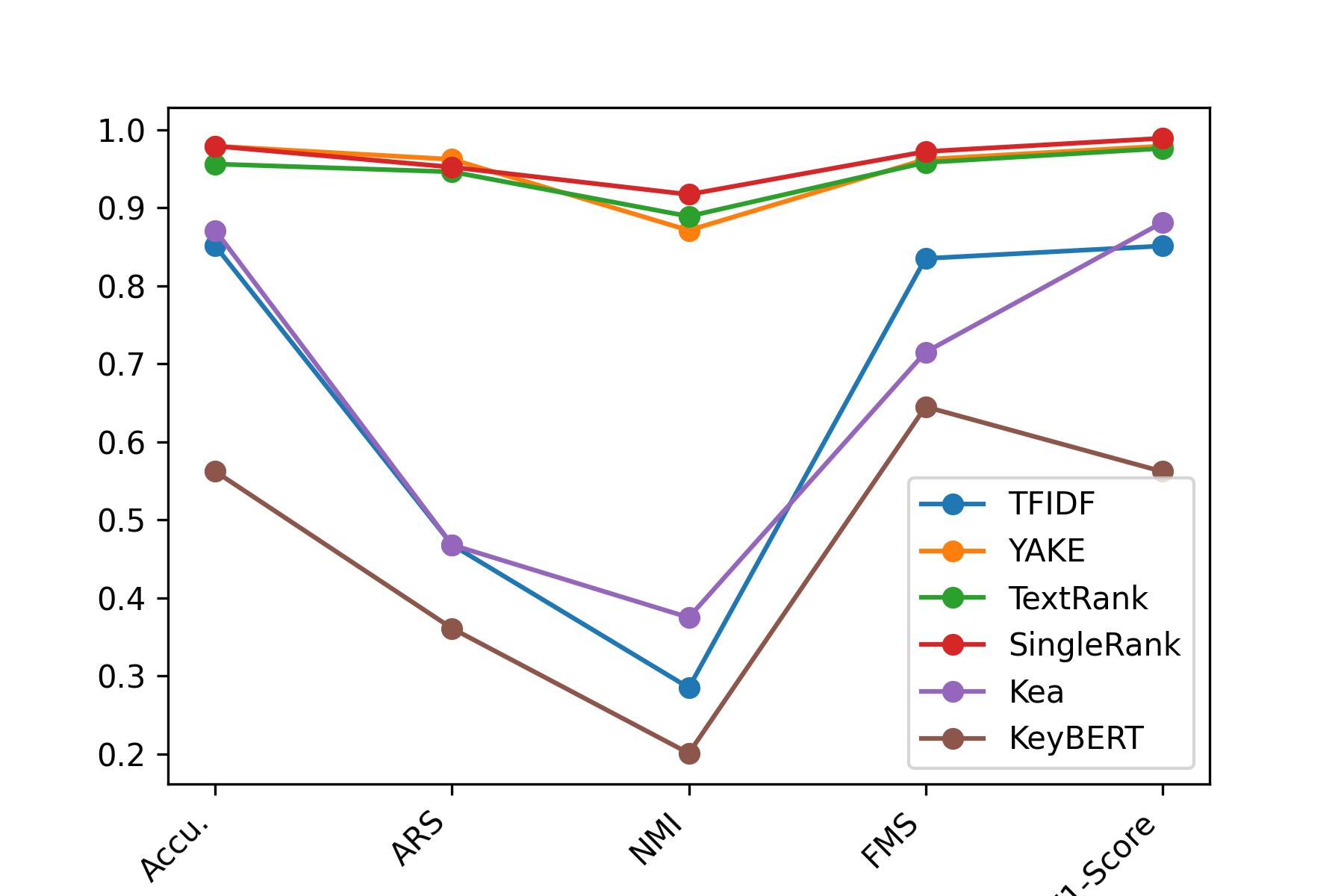}
        \caption{DS-2}
        \label{fig:img2}
    \end{subfigure}
    
    \vspace{-0.1cm}
    
    \begin{subfigure}[b]{0.45\textwidth}
        \centering
        \includegraphics[width=\textwidth]{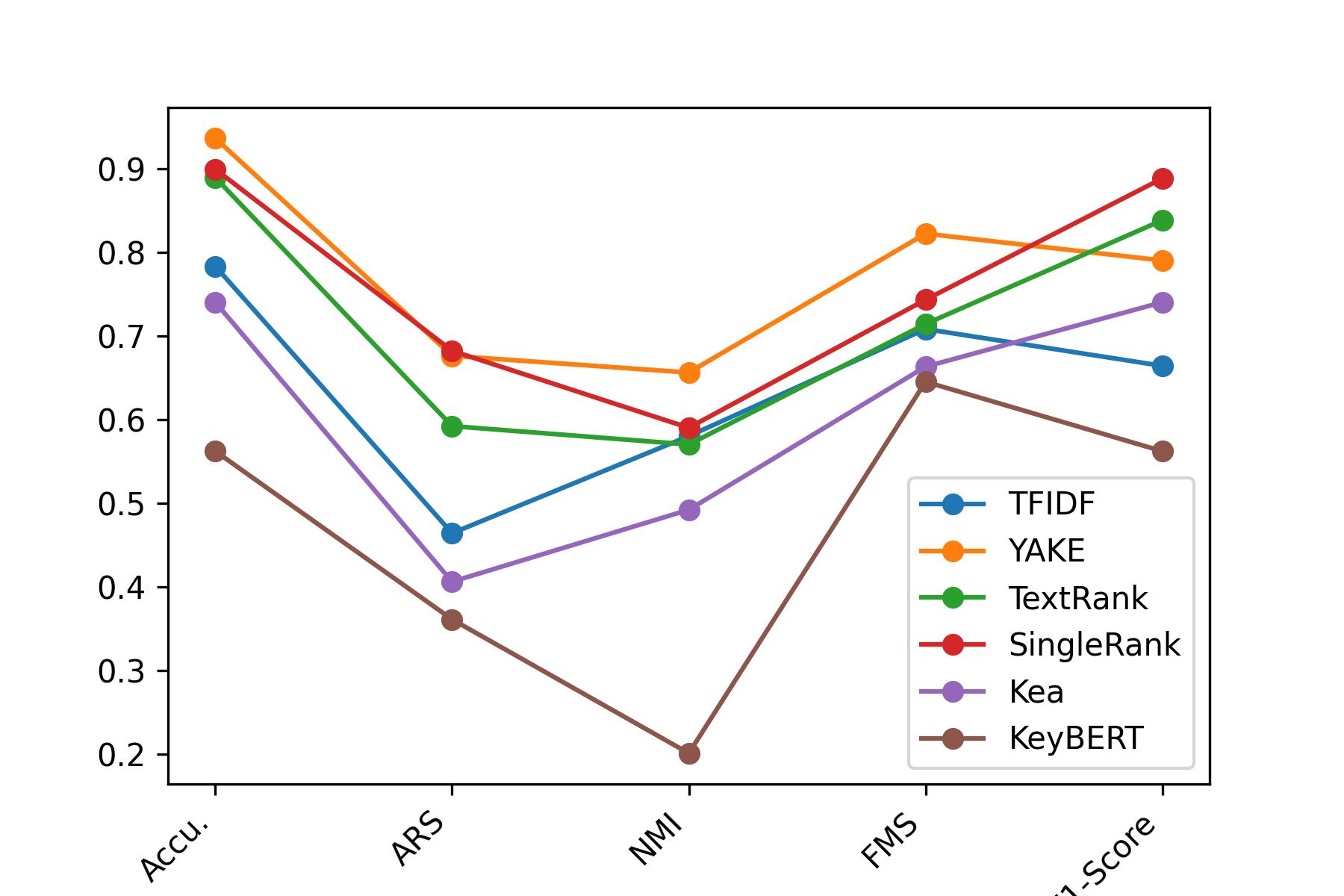}
        \caption{DS-3}
        \label{fig:img3}
    \end{subfigure}
    \hfill
    \begin{subfigure}[b]{0.45\textwidth}
        \centering
        \includegraphics[width=\textwidth]{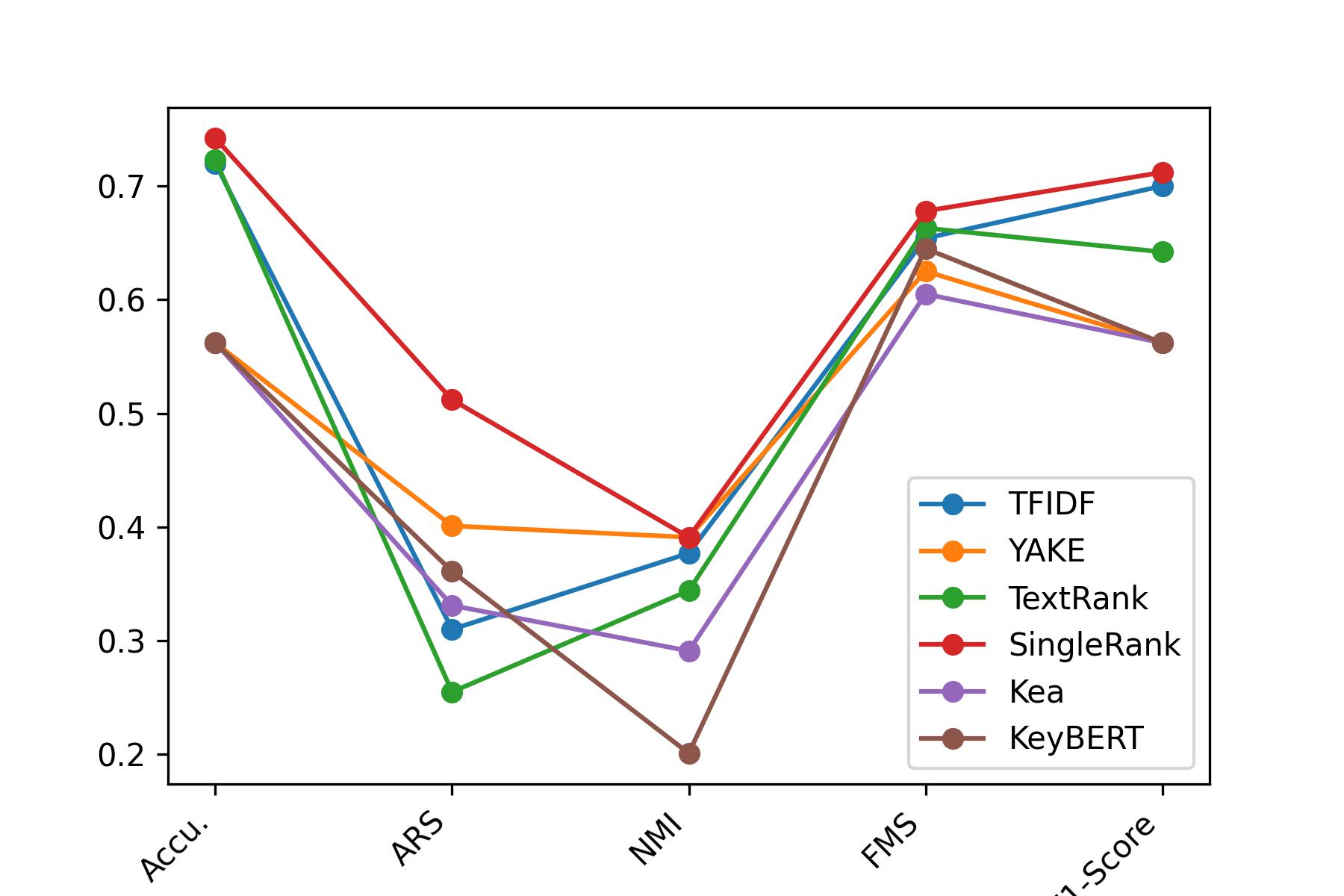}
         \caption{DS-4}
        \label{fig:img4}
    \end{subfigure}
    \caption{A performance analysis of keyword extraction techniques for effective label detection for unknown classes in OpenCML}
    \label{fig:four_images}
\end{figure*}


\begin{table*}[htb!]
\scriptsize
\centering
\caption{Comparison of OpenCML with other recent proposed works of continual machine learning for natural language processing, none of them offers continual learning with open-text classification }
\label{theoryCom}
\begin{tabular}{p{1cm}p{2cm}p{2cm}p{2cm}p{2cm}p{2cm}} \\ \hline
Authors            & Dataset                         & OTC                                  & \multicolumn{3}{c}{CML}                                                               \\ \hline
                   &                                 &                                      & Task              & Number of increments/tasks & Proposed result                      \\ \hline
~\cite{huang2021continual} & bpedia, yahoo, ag, amazon, yelp & NA                                    & TIL  & 5 task                     & Avg  Acc 73.19                       \\
~\cite{ke2021continual}    & Amazon Revies                   & NA                                    & TIL  & 24 task                    & Avg  Acc 0.8524                      \\
~\cite{ermis2022memory}    & Arxiv Papers, Reuters, Wiki-30K & NA                                    & CIL & 5 task                     & Avg Acc 0.88                         \\
                   & twitter data                    & NA                                   & CIL & 5 rounds                   & Avg   Acc 76.8                       \\
\textit{OpenCML}           & DS1, DS2 ,DS3, DS4              & Avg Acc 0.902 \& F1-score 0.831 & CIL & 4 rounds                   & Avg Acc 0.598 \& F1-score 0.709 \\ \hline
\end{tabular}
\end{table*}

\section{Conclusion and Future Work}
 OpenCML, which can discover and identify unknown novel classes, is promising for applications in different domains, including computer vision, natural language processing, and speech recognition. It enables the system to continuously learn from new knowledge without forgetting prior knowledge. The model's capacity to discover new classes without retraining is especially helpful in dynamic environments where new data arrive frequently. Nonetheless, some constraints and challenges still need to be handled. 
One area for future research and essential direction is the integration of knowledge transfer and reinforcement learning into the incremental learning framework. Moreover, the interpretability and explainability of the models should also be considered to secure clarity and accountability. Overall, OpenCML has significant potential and can play an increasingly significant part in future machine learning and AI applications.

\backmatter

\bmhead{Acknowledgements}

Not applicable

\section*{Declarations}

\begin{itemize}
\item Funding: Not applicable
\item Conflict of interest/Competing interests: There is a conflict of Interest 
\item Ethics approval and consent to participate: Not applicable
\item Consent for publication:  Yes
\item Data availability : Not applicable
\item Materials availability:  Not applicable
\item Code availability : Yes
\item Author contribution:  All authors contributed equally to this work
\end{itemize}
\bibliography{Ref}

@article{thrun1995learning,
  title={Is learning the n-th thing any easier than learning the first?},
  author={Thrun, Sebastian},
  journal={Advances in neural information processing systems},
  volume={8},
  year={1995}
}

@incollection{mccloskey1989catastrophic,
  title={Catastrophic interference in connectionist networks: The sequential learning problem},
  author={McCloskey, Michael and Cohen, Neal J},
  booktitle={Psychology of learning and motivation},
  volume={24},
  pages={109--165},
  year={1989},
  publisher={Elsevier}
}

@article{kirkpatrick2017overcoming,
  title={Overcoming catastrophic forgetting in neural networks},
  author={Kirkpatrick, James and Pascanu, Razvan and Rabinowitz, Neil and Veness, Joel and Desjardins, Guillaume and Rusu, Andrei A and Milan, Kieran and Quan, John and Ramalho, Tiago and Grabska-Barwinska, Agnieszka and others},
  journal={Proceedings of the national academy of sciences},
  volume={114},
  number={13},
  pages={3521--3526},
  year={2017},
  publisher={National Acad Sciences}
}

@article{chen2018lifelong,
  title={Lifelong machine learning},
  author={Chen, Zhiyuan and Liu, Bing},
  journal={Synthesis Lectures on Artificial Intelligence and Machine Learning},
  volume={12},
  number={3},
  pages={1--207},
  year={2018},
  publisher={Morgan \& Claypool Publishers}
}

@article{bottou2014machine,
  title={From machine learning to machine reasoning: An essay},
  author={Bottou, L{\'e}on},
  journal={Machine learning},
  volume={94},
  pages={133--149},
  year={2014},
  publisher={Springer}
}

@article{burkart2021survey,
  title={A survey on the explainability of supervised machine learning},
  author={Burkart, Nadia and Huber, Marco F},
  journal={Journal of Artificial Intelligence Research},
  volume={70},
  pages={245--317},
  year={2021}
}

@article{kotsiantis2007supervised,
  title={Supervised machine learning: A review of classification techniques},
  author={Kotsiantis, Sotiris B and Zaharakis, Ioannis and Pintelas, P and others},
  journal={Emerging artificial intelligence applications in computer engineering},
  volume={160},
  number={1},
  pages={3--24},
  year={2007},
  publisher={Amsterdam}
}

@inproceedings{bendale2016towards,
  title={Towards open set deep networks},
  author={Bendale, Abhijit and Boult, Terrance E},
  booktitle={Proceedings of the IEEE conference on computer vision and pattern recognition},
  pages={1563--1572},
  year={2016}
}

@inproceedings{shu2017doc,
  title={DOC: Deep Open Classification of Text Documents},
  author={Shu, Lei and Xu, Hu and Liu, Bing},
  booktitle={Proceedings of the 2017 Conference on Empirical Methods in Natural Language Processing},
  pages={2911--2916},
  year={2017}
}

@incollection{auer2007dbpedia,
  title={Dbpedia: A nucleus for a web of open data},
  author={Auer, S{\"o}ren and Bizer, Christian and Kobilarov, Georgi and Lehmann, Jens and Cyganiak, Richard and Ives, Zachary},
  booktitle={The semantic web},
  pages={722--735},
  year={2007},
  publisher={Springer}
}

@inproceedings{fei2016breaking,
  title={Breaking the closed world assumption in text classification},
  author={Fei, Geli and Liu, Bing},
  booktitle={Proceedings of the 2016 Conference of the North American Chapter of the Association for Computational Linguistics: Human Language Technologies},
  pages={506--514},
  year={2016}
}

@inproceedings{prakhya2017open,
  title={Open set text classification using convolutional neural networks},
  author={Prakhya, Sridhama and Venkataram, Vinodini and Kalita, Jugal},
  booktitle={International Conference on Natural Language Processing, 2017},
  year={2017}
}

@inproceedings{lin2019deep,
  title={Deep Unknown Intent Detection with Margin Loss},
  author={Lin, Ting-En and Xu, Hua},
  booktitle={Proceedings of the 57th Annual Meeting of the Association for Computational Linguistics},
  pages={5491--5496},
  year={2019}
}

@article{hendrycks2016baseline,
  title={A baseline for detecting misclassified and out-of-distribution examples in neural networks},
  author={Hendrycks, Dan and Gimpel, Kevin},
  journal={arXiv preprint arXiv:1610.02136},
  year={2016}
}

@article{vedula2019towards,
  title={Towards open intent discovery for conversational text},
  author={Vedula, Nikhita and Lipka, Nedim and Maneriker, Pranav and Parthasarathy, Srinivasan},
  journal={arXiv preprint arXiv:1904.08524},
  year={2019}
}

@article{lin2019post,
  title={A post-processing method for detecting unknown intent of dialogue system via pre-trained deep neural network classifier},
  author={Lin, Ting-En and Xu, Hua},
  journal={Knowledge-Based Systems},
  volume={186},
  pages={104979},
  year={2019}
}

@article{vedula2020automatic,
  title={Automatic discovery of novel intents \& domains from text utterances},
  author={Vedula, Nikhita and Gupta, Rahul and Alok, Aman and Sridhar, Mukund},
  journal={arXiv preprint arXiv:2006.01208},
  year={2020}
}

@inproceedings{serra2018overcoming,
  title={Overcoming catastrophic forgetting with hard attention to the task},
  author={Serra, Joan and Suris, Didac and Miron, Marius and Karatzoglou, Alexandros},
  booktitle={International Conference on Machine Learning},
  pages={4548--4557},
  year={2018},
  organization={PMLR}
}

@article{gallardo2021self,
  title={Self-supervised training enhances online continual learning},
  author={Gallardo, Jhair and Hayes, Tyler L and Kanan, Christopher},
  journal={arXiv preprint arXiv:2103.14010},
  year={2021}
}

@article{javed2019meta,
  title={Meta-learning representations for continual learning},
  author={Javed, Khurram and White, Martha},
  journal={Advances in Neural Information Processing Systems},
  volume={32},
  year={2019}
}

@article{mehta2021empirical,
  title={An empirical investigation of the role of pre-training in lifelong learning},
  author={Mehta, Sanket Vaibhav and Patil, Darshan and Chandar, Sarath and Strubell, Emma},
  journal={arXiv preprint arXiv:2112.09153},
  year={2021}
}

@inproceedings{wu2022class,
  title={Class-Incremental Learning with Strong Pre-trained Models},
  author={Wu, Tz-Ying and Swaminathan, Gurumurthy and Li, Zhizhong and Ravichandran, Avinash and Vasconcelos, Nuno and Bhotika, Rahul and Soatto, Stefano},
  booktitle={Proceedings of the IEEE/CVF Conference on Computer Vision and Pattern Recognition},
  pages={9601--9610},
  year={2022}
}

@inproceedings{madaan2021representational,
  title={Representational continuity for unsupervised continual learning},
  author={Madaan, Divyam and Yoon, Jaehong and Li, Yuanchun and Liu, Yunxin and Hwang, Sung Ju},
  booktitle={International Conference on Learning Representations},
  year={2021}
}

@inproceedings{purushwalkam2022challenges,
  title={The challenges of continuous self-supervised learning},
  author={Purushwalkam, Senthil and Morgado, Pedro and Gupta, Abhinav},
  booktitle={Computer Vision--ECCV 2022: 17th European Conference, Tel Aviv, Israel, October 23--27, 2022, Proceedings, Part XXVI},
  pages={702--721},
  year={2022},
  organization={Springer}
}

@inproceedings{fini2022self,
  title={Self-supervised models are continual learners},
  author={Fini, Enrico and Da Costa, Victor G Turrisi and Alameda-Pineda, Xavier and Ricci, Elisa and Alahari, Karteek and Mairal, Julien},
  booktitle={Proceedings of the IEEE/CVF Conference on Computer Vision and Pattern Recognition},
  pages={9621--9630},
  year={2022}
}

@inproceedings{cha2021co2l,
  title={Co2l: Contrastive continual learning},
  author={Cha, Hyuntak and Lee, Jaeho and Shin, Jinwoo},
  booktitle={Proceedings of the IEEE/CVF International conference on computer vision},
  pages={9516--9525},
  year={2021}
}

@inproceedings{riemer2018learning,
  title={Learning to Learn without Forgetting by Maximizing Transfer and Minimizing Interference},
  author={Riemer, Matthew and Cases, Ignacio and Ajemian, Robert and Liu, Miao and Rish, Irina and Tu, Yuhai and Tesauro, Gerald},
  booktitle={International Conference on Learning Representations},
  year={2018}
}

@article{chaudhry2019tiny,
  title={On tiny episodic memories in continual learning},
  author={Chaudhry, Arslan and Rohrbach, Marcus and Elhoseiny, Mohamed and Ajanthan, Thalaiyasingam and Dokania, Puneet K and Torr, Philip HS and Ranzato, Marc'Aurelio},
  journal={arXiv preprint arXiv:1902.10486},
  year={2019}
}

@inproceedings{rebuffi2017icarl,
  title={icarl: Incremental classifier and representation learning},
  author={Rebuffi, Sylvestre-Alvise and Kolesnikov, Alexander and Sperl, Georg and Lampert, Christoph H},
  booktitle={Proceedings of the IEEE Conference on Computer Vision and Pattern Recognition},
  pages={2001--2010},
  year={2017}
}

@inproceedings{caccia2020online,
  title={Online learned continual compression with adaptive quantization modules},
  author={Caccia, Lucas and Belilovsky, Eugene and Caccia, Massimo and Pineau, Joelle},
  booktitle={International Conference on Machine Learning},
  pages={1240--1250},
  year={2020},
  organization={PMLR}
}

@inproceedings{belouadah2019il2m,
  title={Il2m: Class incremental learning with dual memory},
  author={Belouadah, Eden and Popescu, Adrian},
  booktitle={Proceedings of the IEEE/CVF International Conference on Computer Vision},
  pages={583--592},
  year={2019}
}

@inproceedings{gong2022continual,
  title={Continual Pre-training of Language Models for Math Problem Understanding with Syntax-Aware Memory Network},
  author={Gong, Zheng and Zhou, Kun and Zhao, Wayne Xin and Sha, Jing and Wang, Shijin and Wen, Ji-Rong},
  booktitle={Proceedings of the 60th Annual Meeting of the Association for Computational Linguistics (Volume 1: Long Papers)},
  pages={5923--5933},
  year={2022}
}

@inproceedings{ebrahimi2020remembering,
  title={Remembering for the Right Reasons: Explanations Reduce Catastrophic Forgetting},
  author={Ebrahimi, Sayna and Petryk, Suzanne and Gokul, Akash and Gan, William and Gonzalez, Joseph E and Rohrbach, Marcus and others},
  booktitle={International Conference on Learning Representations},
  year={2020}
}

@inproceedings{saha2020gradient,
  title={Gradient Projection Memory for Continual Learning},
  author={Saha, Gobinda and Garg, Isha and Roy, Kaushik},
  booktitle={International Conference on Learning Representations},
  year={2020}
}

@inproceedings{schwarz2018progress,
  title={Progress \& compress: A scalable framework for continual learning},
  author={Schwarz, Jonathan and Czarnecki, Wojciech and Luketina, Jelena and Grabska-Barwinska, Agnieszka and Teh, Yee Whye and Pascanu, Razvan and Hadsell, Raia},
  booktitle={International conference on machine learning},
  pages={4528--4537},
  year={2018},
  organization={PMLR}
}

@inproceedings{zenke2017continual_si,
  title={Continual learning through synaptic intelligence},
  author={Zenke, Friedemann and Poole, Ben and Ganguli, Surya},
  booktitle={International Conference on Machine Learning},
  pages={3987--3995},
  year={2017},
  organization={PMLR}
}

@inproceedings{aljundi2018memory_mas,
  title={Memory aware synapses: Learning what (not) to forget},
  author={Aljundi, Rahaf and Babiloni, Francesca and Elhoseiny, Mohamed and Rohrbach, Marcus and Tuytelaars, Tinne},
  booktitle={Proceedings of the European Conference on Computer Vision },
  pages={139--154},
  year={2018}
}

@inproceedings{chaudhry2018riemannian_rwalk,
  title={Riemannian walk for incremental learning: Understanding forgetting and intransigence},
  author={Chaudhry, Arslan and Dokania, Puneet K and Ajanthan, Thalaiyasingam and Torr, Philip HS},
  booktitle={Proceedings of the European Conference on Computer Vision },
  pages={532--547},
  year={2018}
}

@inproceedings{benzing2022unifying,
  title={Unifying Importance Based Regularisation Methods for Continual Learning},
  author={Benzing, Frederik},
  booktitle={International Conference on Artificial Intelligence and Statistics},
  pages={2372--2396},
  year={2022},
  organization={PMLR}
}

@article{lopez2017gradient_gem,
  title={Gradient episodic memory for continual learning},
  author={Lopez-Paz, David and Ranzato, Marc'Aurelio},
  journal={Advances in Neural Information Processing Systems},
  volume={30},
  year={2017}
}

@inproceedings{chaudhry2018efficient_agem,
  title={Efficient Lifelong Learning with A-GEM},
  author={Chaudhry, Arslan and Ranzato, Marc’Aurelio and Rohrbach, Marcus and Elhoseiny, Mohamed},
  booktitle={International Conference on Learning Representations},
  year={2018}
}

@inproceedings{tang2021layerwise,
  title={Layerwise optimization by gradient decomposition for continual learning},
  author={Tang, Shixiang and Chen, Dapeng and Zhu, Jinguo and Yu, Shijie and Ouyang, Wanli},
  booktitle={Proceedings of the IEEE/CVF Conference on Computer Vision and Pattern Recognition},
  pages={9634--9643},
  year={2021}
}

@article{robbins1951stochastic,
  title={A stochastic approximation method},
  author={Robbins, Herbert and Monro, Sutton},
  journal={The annals of mathematical statistics},
  pages={400--407},
  year={1951},
  publisher={JSTOR}
}

@article{kiefer1952stochastic,
  title={Stochastic estimation of the maximum of a regression function},
  author={Kiefer, Jack and Wolfowitz, Jacob},
  journal={The Annals of Mathematical Statistics},
  pages={462--466},
  year={1952},
  publisher={JSTOR}
}

@article{bottou2018optimization,
  title={Optimization methods for large-scale machine learning},
  author={Bottou, L{\'e}on and Curtis, Frank E and Nocedal, Jorge},
  journal={SIAM review},
  volume={60},
  number={2},
  pages={223--311},
  year={2018},
  publisher={SIAM}
}

@inproceedings{Robins93a,
  author    = {Anthony V. Robins},
  title     = {Catastrophic forgetting in neural networks: the role of rehearsal
               mechanisms},
  booktitle = {International Two-Stream Conference on Artificial Neural Networks and Expert Systems, {ANNES}},
  year      = {1993}
}

@article{Robins95,
  author  = {Anthony V. Robins},
  title   = {Catastrophic Forgetting, Rehearsal and Pseudorehearsal},
  journal = {Connect. Sci.},
  year    = {1995}
}

@article{FrenchC02,
  author  = {Robert M. French and
               Nick Chater},
  title   = {Using Noise to Compute Error Surfaces in Connectionist Networks: {A}
               Novel Means of Reducing Catastrophic Forgetting},
  journal = {Neural Comput.},
  year    = {2002}
}

@inproceedings{castro2018end,
  title={End-to-end incremental learning},
  author={Castro, Francisco M and Mar{\'\i}n-Jim{\'e}nez, Manuel J and Guil, Nicol{\'a}s and Schmid, Cordelia and Alahari, Karteek},
  booktitle={Proceedings of the European conference on computer vision (ECCV)},
  pages={233--248},
  year={2018}
}

@inproceedings{welling2009herding,
  title={Herding dynamical weights to learn},
  author={Welling, Max},
  booktitle={Proceedings of the 26th Annual International Conference on Machine Learning},
  pages={1121--1128},
  year={2009}
}

@article{hinton2015distilling,
  title={Distilling the knowledge in a neural network},
  author={Hinton, Geoffrey and Vinyals, Oriol and Dean, Jeff},
  journal={arXiv preprint arXiv:1503.02531},
  year={2015}
}

@inproceedings{larson-etal-2019-evaluation,
    title = "An Evaluation Dataset for Intent Classification and Out-of-Scope Prediction",
    author = "Larson, Stefan  and
      Mahendran, Anish  and
      Peper, Joseph J.  and
      Clarke, Christopher  and
      Lee, Andrew  and
      Hill, Parker  and
      Kummerfeld, Jonathan K.  and
      Leach, Kevin  and
      Laurenzano, Michael A.  and
      Tang, Lingjia  and
      Mars, Jason",
    booktitle = "International Joint Conference on Natural Language Processing",
    pages = "1311--1316",
    year = "2019", 
}

@inproceedings{casanueva2020efficient,
title={Efficient intent detection with dual sentence encoders},
author={Casanueva, Iker and Tem\v{c}inas, Tomas and Gerz, David and Henderson, Matthew and Vuli'{c}, Ivan},
booktitle={proceedings of the 28th International Conference on Computational Linguistics},
pages={38--45},
year={2020}
}

@inproceedings{zhang2021deep,
  title={Deep open intent classification with adaptive decision boundary},
  author={Zhang, Hanlei and Xu, Hua and Lin, Ting-En},
  booktitle={proceedings of the AAAI Conference on Artificial Intelligence},
  volume={35},
  number={16},
  pages={14374--14382},
  year={2021}
}

@inproceedings{yan2020unknown,
  title={Unknown intent detection using Gaussian mixture model with an application to zero-shot intent classification},
  author={Yan, Guangfeng and Fan, Lu and Li, Qimai and Liu, Han and Zhang, Xiaotong and Wu, Xiao-Ming and Lam, Albert YS},
  booktitle={proceedings of the 58th annual meeting of the association for computational linguistics},
  pages={1050--1060},
  year={2020}
}

@inproceedings{zhou2022knn,
  title={KNN-contrastive learning for out-of-domain intent classification},
  author={Zhou, Yunhua and Liu, Peiju and Qiu, Xipeng},
  booktitle={proceedings of the 60th Annual Meeting of the Association for Computational Linguistics (Volume 1: Long Papers)},
  pages={5129--5141},
  year={2022}
}

@inproceedings{zeng2021modeling,
  title={Modeling Discriminative Representations for Out-of-Domain Detection with Supervised Contrastive Learning},
  author={Zeng, Zhiyuan and He, Keqing and Yan, Yuanmeng and Liu, Zijun and Wu, Yanan and Xu, Hong and Jiang, Huixing and Xu, Weiran},
  booktitle={proceedings of the 59th Annual Meeting of the Association for Computational Linguistics and the 11th International Joint Conference on Natural Language Processing (Volume 2: Short Papers)},
  pages={870--878},
  year={2021}
}

@article{huang2021continual,
  title={Continual learning for text classification with information disentanglement based regularization},
  author={Huang, Yufan and Zhang, Yanzhe and Chen, Jiaao and Wang, Xuezhi and Yang, Diyi},
  journal={arXiv preprint arXiv:2104.05489},
  year={2021}
}

@inproceedings{ke2021continual,
  title={Continual learning with knowledge transfer for sentiment classification},
  author={Ke, Zixuan and Liu, Bing and Wang, Hao and Shu, Lei},
  booktitle={Machine Learning and Knowledge Discovery in Databases: European Conference, ECML PKDD 2020, Ghent, Belgium, September 14--18, 2020, Proceedings, Part III},
  pages={683--698},
  year={2021},
  organization={Springer}
}

@article{ermis2022memory,
  title={Memory efficient continual learning with transformers},
  author={Ermis, Beyza and Zappella, Giovanni and Wistuba, Martin and Rawal, Aditya and Archambeau, Cedric},
  journal={Advances in Neural Information Processing Systems},
  volume={35},
  pages={10629--10642},
  year={2022}
}

@inproceedings{xu2015short,
  title={Short text clustering via convolutional neural networks},
  author={Xu, Jiaming and Wang, Peng and Tian, Guanhua and Xu, Bo and Zhao, Jun and Wang, Fangyuan and Hao, Hongwei},
  booktitle={Proceedings of the 1st Workshop on Vector Space Modeling for Natural Language Processing},
  pages={62--69},
  year={2015}
}





\end{document}